\documentclass{article} 
\usepackage{iclr2024_conference,times}

\usepackage{booktabs,multirow,longtable,makecell} 
\usepackage{amsmath,amsfonts,bm}

\def\eqref#1{equation~\ref{#1}}

\def\1{\bm{1}}

\DeclareMathAlphabet{\mathsfit}{\encodingdefault}{\sfdefault}{m}{sl}
\SetMathAlphabet{\mathsfit}{bold}{\encodingdefault}{\sfdefault}{bx}{n}

\usepackage{hyperref}
\usepackage{url}
\usepackage{xspace}
\usepackage{enumitem}
\usepackage{graphicx}
\usepackage{wrapfig}
\usepackage{fontawesome5}
\usepackage{subcaption}
\usepackage{multirow}
\usepackage{tcolorbox}
\usepackage{wrapfig}
\usepackage{listings}
\usepackage{xcolor}
\usepackage[capitalize]{cleveref}
\usepackage{marvosym}
\newcommand{\framework}{IoA\xspace}

\colorlet{punct}{red!60!black}
\definecolor{background}{HTML}{EEEEEE}
\definecolor{delim}{RGB}{20,105,176}
\colorlet{numb}{magenta!60!black}
\lstdefinelanguage{json}{
    basicstyle=\normalfont\ttfamily,
    numbers=left,
    numberstyle=\scriptsize,
    stepnumber=1,
    numbersep=8pt,
    showstringspaces=false,
    breaklines=true,
    frame=lines,
    backgroundcolor=\color{background},
    literate=
     *{0}{{{\color{numb}0}}}{1}
      {1}{{{\color{numb}1}}}{1}
      {2}{{{\color{numb}2}}}{1}
      {3}{{{\color{numb}3}}}{1}
      {4}{{{\color{numb}4}}}{1}
      {5}{{{\color{numb}5}}}{1}
      {6}{{{\color{numb}6}}}{1}
      {7}{{{\color{numb}7}}}{1}
      {8}{{{\color{numb}8}}}{1}
      {9}{{{\color{numb}9}}}{1}
      {:}{{{\color{punct}{:}}}}{1}
      {,}{{{\color{punct}{,}}}}{1}
      {\{}{{{\color{delim}{\{}}}}{1}
      {\}}{{{\color{delim}{\}}}}}{1}
      {[}{{{\color{delim}{[}}}}{1}
      {]}{{{\color{delim}{]}}}}{1},
}

\title{\includegraphics[width=1.5em]{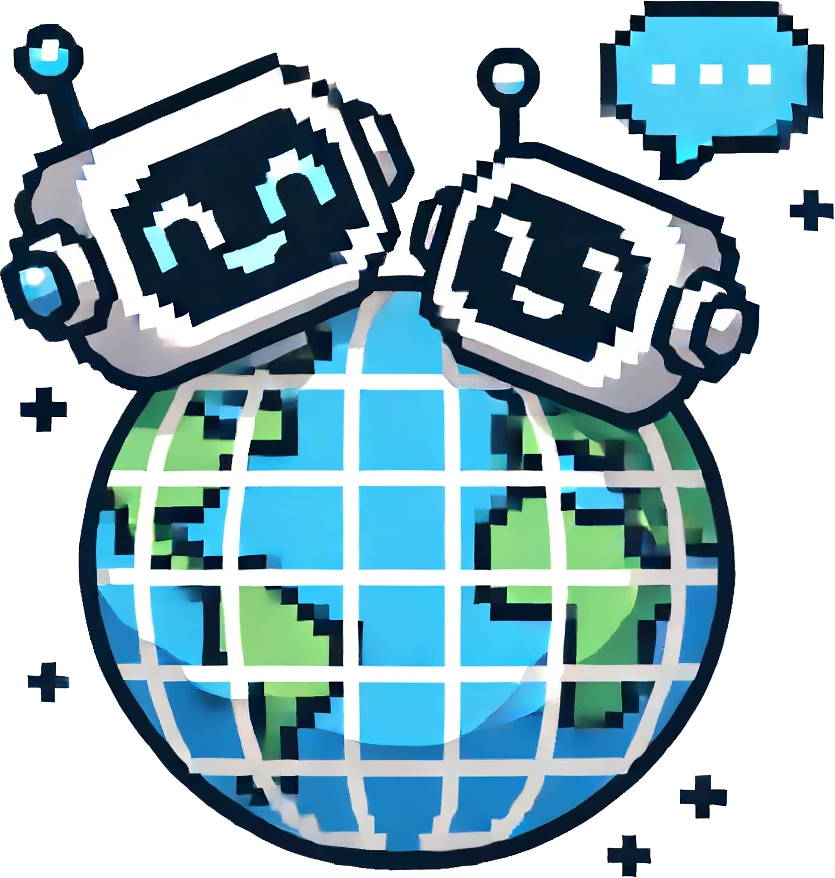}Internet of Agents: Weaving a Web of Heterogeneous Agents for Collaborative Intelligence}

\author{Weize Chen$^1\thanks{Equal Contribution. \quad\Letter \ Corresponding author.}$\hspace{0.5em}, Ziming You$^{2*}$, Ran Li$^{1*}$, Yitong Guan$^{2*}$, Chen Qian$^1$, Chenyang Zhao$^1$\\\textbf{Cheng Yang$^3$, Ruobing Xie$^4$, Zhiyuan Liu$^1$\textsuperscript{\Letter}, Maosong Sun$^1$}\\
$^1$ Tsinghua University,
$^2$ Peking University\\
$^3$ Beijing University of Posts and Telecommunications,
$^4$ Tencent\\
\texttt{chenwz21@mails.tsinghua.edu.cn},\quad \texttt{ran.li572482@gmail.com}\\
\texttt{\{zimingyou, 2101210206\}@stu.pku.edu.cn}\\
\texttt{liuzy@tsinghua.edu.cn}
}
\iclrfinalcopy 
\begin{document}

\maketitle

\begin{abstract}
The rapid advancement of large language models (LLMs) has paved the way for the development of highly capable autonomous agents. However, existing multi-agent frameworks often struggle with integrating diverse capable third-party agents due to reliance on agents defined within their own ecosystems. They also face challenges in simulating distributed environments, as most frameworks are limited to single-device setups. Furthermore, these frameworks often rely on hard-coded communication pipelines, limiting their adaptability to dynamic task requirements. Inspired by the concept of the Internet, we propose the Internet of Agents (IoA), a novel framework that addresses these limitations by providing a flexible and scalable platform for LLM-based multi-agent collaboration. \framework introduces an agent integration protocol, an instant-messaging-like architecture design, and dynamic mechanisms for agent teaming and conversation flow control. Through extensive experiments on general assistant tasks, embodied AI tasks, and retrieval-augmented generation benchmarks, we demonstrate that \framework consistently outperforms state-of-the-art baselines, showcasing its ability to facilitate effective collaboration among heterogeneous agents. \framework represents a step towards linking diverse agents in an Internet-like environment, where agents can seamlessly collaborate to achieve greater intelligence and capabilities. Our codebase has been released at \url{https://github.com/OpenBMB/IoA}.
\end{abstract}

\section{Introduction}
\label{sec:introduction}

The Internet has revolutionized the way people collaborate and share knowledge, connecting individuals with diverse skills and backgrounds from all around the world. This global network has enabled the creation of remarkable collaborative projects, such as Wikipedia\footnote{\url{https://www.wikipedia.org/}} and the development of the Linux operating system\footnote{\url{https://www.linux.org/}}, which would have been impossible for any single person to achieve. The Internet has greatly facilitated collaboration among people, making the impossible possible and pushing the boundaries of human achievement.

The success of the Internet in enabling human collaboration raises an intriguing question: can we create a similar platform to facilitate collaboration among autonomous agents? With the rapid advancements in LLMs~\citep{DBLP:journals/corr/abs-2303-08774, DBLP:journals/corr/abs-2403-05530}, we now have autonomous agents capable of achieving near-human performance on a wide range of tasks. These LLM-based agents have demonstrated the ability to break down complex tasks into executable steps, leverage various tools, and learn from feedback and experience~\citep{DBLP:journals/corr/abs-2307-16789, DBLP:journals/corr/abs-2309-10691, DBLP:conf/nips/ShinnCGNY23, DBLP:journals/corr/abs-2312-17025}. As the capabilities of these agents continue to grow, and with an increasing number of third-party agents with diverse skills consistently emerging~\citep{Chase_LangChain_2022,xagent2023,Significant_Gravitas_AutoGPT,OpenInterpreter}, it is crucial to explore how we can effectively and efficiently orchestrate their collaboration, just as the Internet has done for humans.

To address this challenge, we propose the concept of the Internet of Agents (\framework), a general framework for agent communication and collaboration inspired by the Internet. \framework aims to address three fundamental limitations of existing multi-agent frameworks~\citep{chen2023agentverse,DBLP:journals/corr/abs-2308-08155,DBLP:journals/corr/abs-2308-00352,DBLP:journals/corr/abs-2307-07924}: (1) \textbf{Ecosystem Isolation}: Most frameworks only consider agents defined within their own ecosystems, potentially blocking the integration of various third-party agents and limiting the diversity of agent capabilities and the platform's generality; (2) \textbf{Single-Device Simulation}: Nearly all multi-agent frameworks simulate multi-agent systems on a single device, which differs significantly from real-world scenarios where agents could be distributed across multiple devices located in different places; (3) \textbf{Rigid Communication and Coordination}: The communication process, agent grouping, and state transitions are mostly hard-coded, whereas in real life, humans decide on teammates based on the task at hand and dynamically switch between discussion and task assignment or execution.

To overcome these limitations, we propose an agent integration protocol that enables different third-party agents running on different devices to be seamlessly integrated into the framework and collaborate effectively. Additionally, we introduce an instant-messaging-app-like framework that facilitates agent discovery and dynamic teaming. By autonomously searching for potential agents capable of handling the tasks at hand, agents can dynamically decide to form different teams and communicate within various group chats. Inspired by Speech Act Theory~\citep{Searle_1969}, and its application in conventional multi-agent system~\citep{DBLP:conf/cikm/FininFMM94,DBLP:journals/expert/LabrouFP99}, within each group chat, we abstract out several conversation states and provide a flexible and general finite-state machine mechanism that allows agents to autonomously decide the state of the conversation, facilitating discussion and sub-task execution.

We demonstrate the effectiveness of \framework through extensive experiments and comparisons with state-of-the-art autonomous agents. By integrating AutoGPT~\citep{Significant_Gravitas_AutoGPT} and Open Interpreter~\citep{OpenInterpreter}, we show that \framework achieves a 66 to 76\% win rate in open-domain task evaluations when compared with these agents individually. Furthermore, with only a few basic ReAct agents integrated, \framework outperforms previous works on the GAIA benchmark~\citep{DBLP:journals/corr/abs-2311-12983}. In the retrieval-augmented generation (RAG) question-answering domain, our framework substantially surpasses existing methods, with a GPT-3.5-based implementation achieving performance close to or even exceeding GPT-4, and effectively surpassing previous multi-agent framework.

The impressive performance of \framework across various domains highlights the potential of this paradigm for autonomous agents. As smaller LLMs continue to advance~\citep{DBLP:journals/corr/abs-2403-08295,DBLP:journals/corr/abs-2404-06395,DBLP:journals/corr/abs-2404-14219}, running agents on personal computer or even mobile device is becoming increasingly feasible. This trend opens up new opportunities for deploying multi-agent systems in real-world scenarios, where agents can be distributed across multiple devices and collaborate to solve complex problems. We believe that by further exploring and refining the \framework paradigm, more sophisticated and adaptable multi-agent systems can be developed, ultimately pushing the boundaries of what autonomous agents can achieve in problem-solving and decision-making.

\section{Framework Design and Key Mechanisms of \framework}
\label{sec:framework-and-mechanisms}

In this section, we present a comprehensive overview of \framework, detailing its architecture and key mechanisms. We will explore how these components work together to enable effective collaboration among autonomous agents, facilitating dynamic team formation, structured communication, and efficient task execution.

\subsection{Overview of \framework}
\label{sec:framework-overview}

\framework is designed as an instant-messaging-app-like platform that enables seamless communication and collaboration among diverse autonomous agents. Inspired by the concept of Internet, \framework addresses three fundamental challenges in multi-agent systems~\citep{chen2023agentverse,DBLP:journals/corr/abs-2308-08155,DBLP:journals/corr/abs-2307-07924}:

\begin{enumerate}[noitemsep,topsep=0pt,parsep=0pt,partopsep=0pt,leftmargin=1.5em]
    \item Distributed agent collaboration: Unlike traditional frameworks that simulate multi-agent systems on a single device, \framework supports agents distributed across multiple devices and locations. (\cref{sec:framework-architecture,sec:framework-key-mech-registration})
    \item Dynamic and adaptive communication: \framework implements mechanisms for autonomous team formation and conversation flow control, allowing agents to adapt their collaboration strategies based on task requirements and ongoing progress. (\cref{sec:framework-key-mech-teamup,sec:framework-key-mech-flow,sec:framework-key-mech-task-assign})
    \item Integration of heterogeneous agents: \framework provides a flexible protocol for integrating various third-party agents, expanding the diversity of agent capabilities within the system. (\cref{sec:framework-protocol})
\end{enumerate}

At its core, \framework consists of two main components: the server and the client. The server acts as a central hub, managing agent registration, discovery, and message routing. It enables agents with varying capabilities to find each other and initiate communication. The client, on the other hand, serves as a wrapper for individual agents, providing them with the necessary communication functionalities and adapting them to the specified protocol.
\framework employs a layered architecture~\citep{DBLP:books/daglib/0000199} for both the server and client components, comprising three layers:
\begin{itemize}[noitemsep,topsep=0pt,parsep=0pt,partopsep=0pt,leftmargin=1.5em]
    \item \textbf{Interaction Layer:} Facilitates team formation and agent communication.
    \item \textbf{Data Layer:} Manages information related to agents, group chats, and tasks.
    \item \textbf{Foundation Layer:} Provides essential infrastructure for agent integration, data management, and network communication.
\end{itemize}
These layers work together to facilitate agent collaboration through the network. In the following subsections, we will go through the \framework's architecture and design.

\begin{figure}
    \centering
    \includegraphics[width=\textwidth]{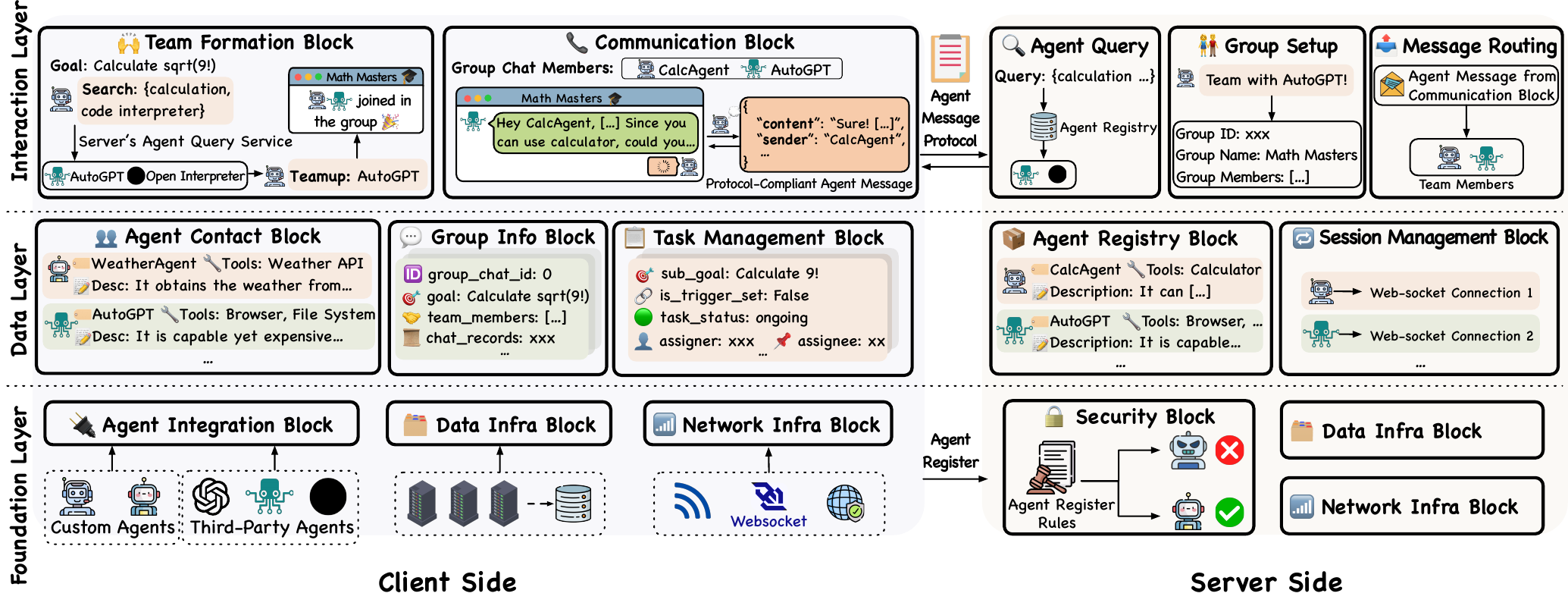}
    \caption{The illustration on the conceptual layered architecture on the design of \framework.}
    \label{fig:layer-architecture}
\end{figure}

\subsection{Architecture of \framework}
\label{sec:framework-architecture}

The layered architecture of \framework is designed to support scalable, flexible, and efficient multi-agent collaboration. This architecture enables a clear separation of concerns and facilitates the integration of diverse agents and functionalities (\cref{fig:layer-architecture}).

\subsubsection{Server Architecture}
\label{sec:framework-architecture-server}

The server acts as the central hub of \framework, facilitating agent discovery, group formation, and message routing. Its architecture consists of three layers:

\textbf{Interaction Layer:} At the top level, the Interaction Layer manages high-level interactions between agents and the system. It encompasses the Agent Query Block for enabling agents to search for other agents based on specific characteristics, the Group Setup Block for facilitating the creation and management of group chats, and the Message Routing Block for ensuring efficient and accurate routing of messages between agents and group chats.

\textbf{Data Layer:} Serving as the information backbone, the Data Layer handles the storage and management of critical system information. The Agent Registry Block maintains a comprehensive database of registered agents, including their capabilities and current status, similar to service discovery in distributed systems~\citep{DBLP:journals/cn/MeshkovaRPM08,Netflix_Eureka}. Meanwhile, the Session Management Block manages active connections and ensures continuous communication between the server and connected clients.

\textbf{Foundation Layer:} Underpinning the entire system, the Foundation Layer provides the essential infrastructure for the server's operations. It encompasses the Data Infrastructure Block for handling data persistence and retrieval, the Network Infrastructure Block for managing network communications, and the Security Block for implementing authentication, authorization, and other security measures to maintain system integrity.

\subsubsection{Client Architecture}
\label{sec:framework-architecture-client}

The client component of \framework serves as a wrapper for individual agents, providing them with the necessary interfaces to communicate within the system. Its architecture mirrors that of the server with three layers:

\textbf{Interaction Layer:} At the forefront of agent operations, the Interaction Layer manages the agent's interactions within the system. The Team Formation Block implements the logic for identifying suitable collaborators and forming teams for the task at hand, similar to coalition formation in conventional multi-agent research~\citep{DBLP:journals/jair/RahwanRJG09}. Complementing this, the Communication Block manages the agent's participation in group chats and handles message processing.

\textbf{Data Layer:} Functioning as the agent's memory, the Data Layer maintains local data relevant to the agent's operations. It includes the Agent Contact Block for storing information about other agents the current agent has interacted with, the Group Info Block for maintaining details about ongoing group chats and collaborations, and the Task Management Block for tracking the status and progress of tasks assigned to the agent.

\textbf{Foundation Layer:} Forming the base of the client architecture, the Foundation Layer provides the basic functionalities for the client's operations. The Agent Integration Block defines the protocols and interfaces for integrating third-party agents into the \framework ecosystem. Alongside this, the Data Infrastructure Block handles local data storage and retrieval, while the Network Infrastructure Block manages network communications with the server.

This layered architecture enables \framework to support a wide range of agent types and collaboration scenarios. By providing a clear separation of concerns and well-defined interfaces between layers, the architecture facilitates the integration of diverse agents and allows for future extensibility. Furthermore, this design supports the key mechanisms of \framework, such as autonomous team formation and conversation flow control, which we will explore in detail in the following subsections.

\subsection{Key Mechanisms}
\label{sec:framework-key-mech}

The effectiveness of \framework relies on several key mechanisms that enable seamless collaboration among diverse agents. These mechanisms work in concert to facilitate agent integration, team formation, task allocation, and structured communication. We detail these critical components in this section.

\subsubsection{Agent Registration and Discovery}
\label{sec:framework-key-mech-registration}

To enable collaboration among distributed agents with heterogeneous architectures, tools, and environments, we propose the agent registration and discovery mechanism. This mechanism forms the foundation for collaborative interactions within \framework, enabling the integration of diverse agents into the system and facilitating their discovery on the online server by other agents for potential collaboration through the network.

\textbf{Agent Registration:} When a new agent joins the \framework, its client wrapper undergoes a registration process with the server. During registration, the agent should provide a comprehensive description of its capabilities, skills, and areas of expertise. This description, denoted as $d_i$ for an agent $c_i$, is stored in the Agent Registry Block of the server's Data Layer. Formally, we represent the set of all registered agents as $\mathcal{C} = \{c_1, c_2, ..., c_n\}$, where each $c_i$ is associated with its description $d_i$.


\textbf{Agent Discovery:} The agent discovery function leverages the information stored in the Agent Registry from the online server to enable agents to find suitable collaborators for specific tasks. When an agent needs to form a team or seek assistance, it can use the $\texttt{search\_client}$ tool provided by the server's Agent Query Block. This tool allows an agent to search for other agents based on desired characteristics or capabilities. Formally, the agent discovery process can be described as follows: Let $\mathcal{L}_d = [l_1, l_2, ..., l_k]$ be a list of desired characteristics generated by an agent seeking collaborators. The $\texttt{search\_client}$ function can be represented as:
$\texttt{search\_client}: \mathcal{L}_d \rightarrow \mathcal{P}(\mathcal{C})$,
where $\mathcal{P}(\mathcal{C})$ denotes the power set of $\mathcal{C}$. The function returns a subset of clients $\mathcal{C}_d \subseteq \mathcal{C}$ whose descriptions $d_j$ match the desired characteristics in $\mathcal{L}_d$.
The matching process between $\mathcal{L}_d$ and $d_j$ can be implemented with various semantic matching techniques~\citep{DBLP:journals/ftir/RobertsonZ09,DBLP:conf/emnlp/KarpukhinOMLWEC20}. It ensures that agents with relevant capabilities can be discovered even if their descriptions do not exactly match the search criteria.


\subsubsection{Autonomous Nested Team Formation}
\label{sec:framework-key-mech-teamup}

The autonomous nested team formation mechanism enables dynamic and flexible combinations of appropriate agents. This mechanism allows agents to form teams adaptively based on task requirements and to create nested sub-teams for complex, multi-faceted tasks.

\textbf{Team Formation Process:} When a client $c_i \in \mathcal{C}$ is assigned a task $t$, it initiates the team formation process. The client has access to two essential tools provided by the server: $\texttt{search\_client}$ and $\texttt{launch\_group\_chat}$. 
The LLM in the client is prompted to decide which tool to call based on the task and the current set of discovered clients. If more collaborators are needed, it calls $\texttt{search\_client}$ with appropriate characteristics. Once suitable collaborators are found, it calls $\texttt{launch\_group\_chat}$ to initiate a new group chat $g \in \mathcal{G}$, where $\mathcal{G}$ is the space of all group chats.

\textbf{Nested Team Structure:} The nested team formation allows for a hierarchical structure of teams and sub-teams. Let $g_0 \in \mathcal{G}$ be the initial group chat for task $t$. During the execution of $t$, if a client $c_i$ is assigned with a sub-task $t_l$ (the task assignment mechanism will be introduced in \cref{sec:framework-key-mech-task-assign}), and it identifies $t_l$ requires additional expertise, $c_i$ is allowed to search for appropriate agents again and initiate a new sub-group chat $g_l \in \mathcal{G}$. This process can continue recursively for the new sub-tasks assigned in $g_l$, forming a tree-like structure of group chats.
Formally, we can define a function $h: \mathcal{G} \rightarrow \mathcal{P}(\mathcal{G})$ that maps a group chat to its set of sub-group chats. The nested structure can be represented as:
$h(g_0) = \{g_1, g_2, ..., g_m\}, \quad h(g_i) = \{g_{i1}, g_{i2}, ..., g_{in}\}$, and so on.

\begin{wrapfigure}{r}{0.5\textwidth}
    \centering
    \vspace{-1.2em}
    \includegraphics[width=0.5\textwidth]{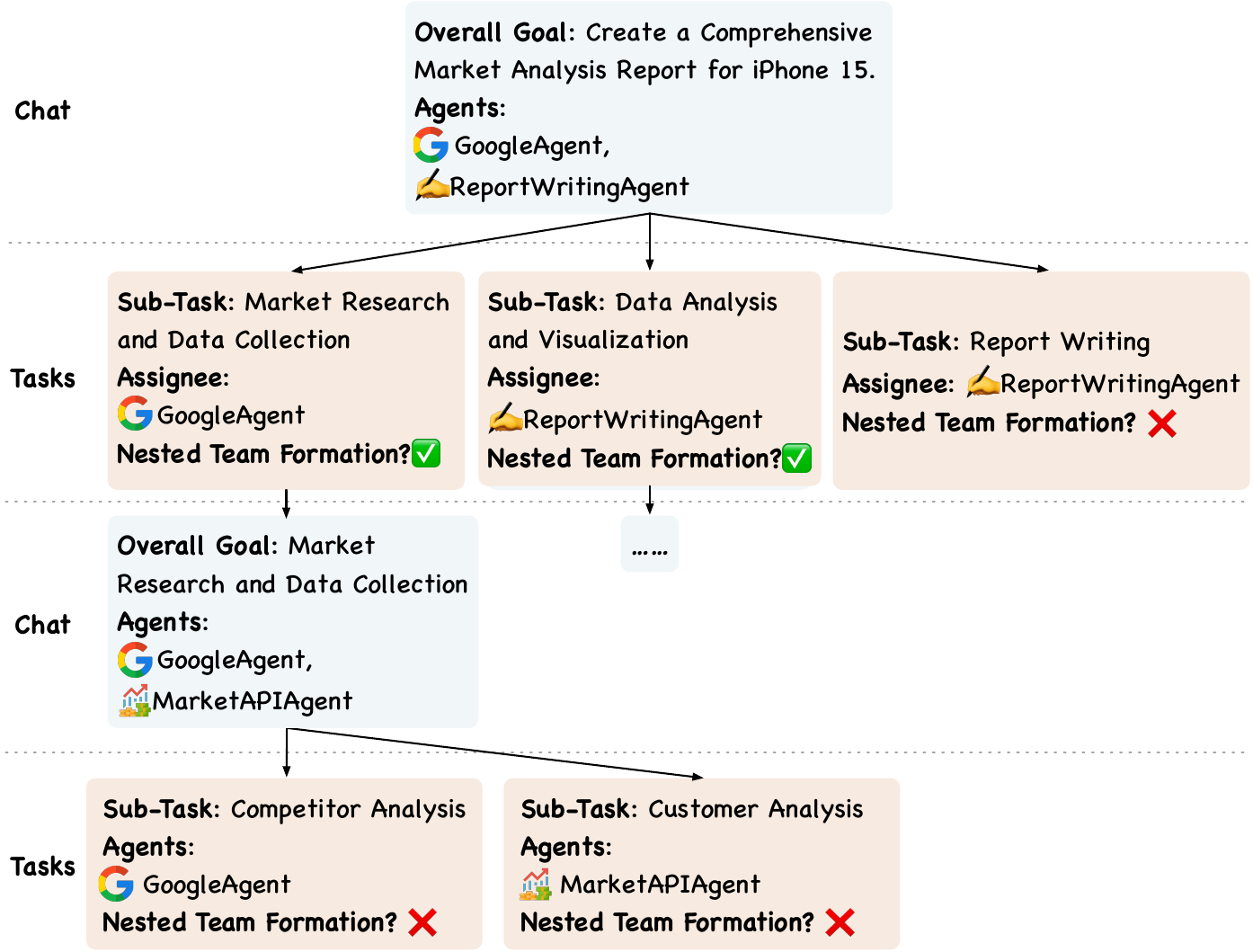}
    \vspace{-1em}
    \caption{An example of nested team formation mechanism. The process is simplified for clarity.}
    \vspace{-1em}
    \label{fig:nested-teamup}
\end{wrapfigure}
\textbf{Communication Complexity:} The nested team formation mechanism helps reduce communication complexity in large agent teams. Assuming fully connected communication within each group, the number of communication channels (connected edges) in a single group with $|g|$ members is $c_\text{full}=\frac{|g|(|g|-1)}{2}$. However, by decomposing a task into sub-tasks and allocating them to sub-group chats, we can reduce the total number of communication channels.
Let $\mathcal{S}(g)$ denote the set of all sub-groups (including $g$ itself) formed for a task initially assigned to group $g$. The total number of communication channels can then be expressed as:
$c_\text{nested}=\sum_{g_i \in \mathcal{S}(g)} \frac{|g_i|(|g_i|-1)}{2}\le c_\text{full}$.

\cref{fig:nested-teamup} illustrates an example of the nested team formation process. In this example, the initial group chat $g_0$ spawns three sub-group chats $g_1$, $g_2$ and $g_3$ for specific sub-tasks during the discussion. $g_1$ further creates two sub-group chats $g_{21}$ and $g_{22}$ for a more specialized sub-task.


\subsubsection{Autonomous Conversation Flow Control}
\label{sec:framework-key-mech-flow}

Effective communication is crucial for successful collaboration among autonomous agents. Inspired by Speech Act Theory~\citep{austin1975things, Searle_1969} and its applications in multi-agent systems~\citep{DBLP:conf/cikm/FininFMM94, DBLP:journals/expert/LabrouFP99}, we introduce an autonomous conversation flow control mechanism in \framework. This mechanism enables agents to coordinate their communication and maintain a structured dialogue, enhancing the efficiency and effectiveness of their collaboration.

\textbf{Sequential Speaking Mechanism:} To manage potential conflicts and ensure clear communication, \framework adopts the most basic sequential speaking mechanism. At any given time, only one agent is permitted to speak, preventing confusion and maintaining a clear order of communication. This approach, while simple, provides a foundation for more sophisticated conversation management when combined with the following dynamic features.

\textbf{Finite State Machine for Group Chat States:} We formalize the conversation flow as a finite state machine $M = (S, \Sigma, \delta, s_0, F)$, where:

\begin{itemize}[noitemsep,topsep=0pt,parsep=0pt,partopsep=0pt,leftmargin=1.5em]
    \item $S = \{s_d, s_s, s_a, s_p, s_c\}$ is the set of states representing discussion, synchronous task assignment, asynchronous task assignment, pause \& trigger, and conclusion, respectively.
    \item $\Sigma$ is the state transition decision space.
    \item $\delta: S \times \Sigma \to S$ is the transition function mapping the current state and the transition decision made by LLMs to the next state.
    \item $s_0 = s_d$ is the initial state, representing the start of the conversation in the discussion phase.
    \item $F = \{s_c\}$ is the set of final states, containing only the conclusion state.
\end{itemize}

\begin{wrapfigure}{r}{0.4\textwidth}
    \centering
    \vspace{-1.3em}
    \includegraphics[width=0.4\textwidth]{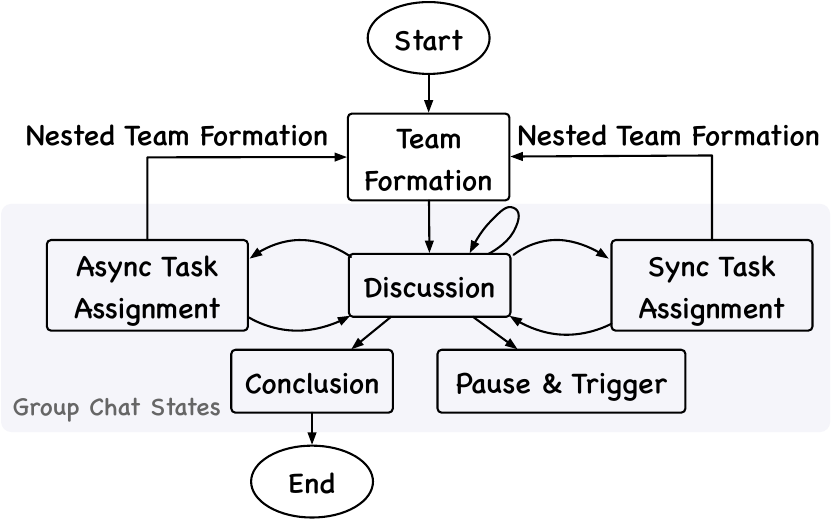}
    \caption{The state transition among different states.}
    \label{fig:flow}
\end{wrapfigure}

Figure~\ref{fig:flow} illustrates the state transitions in the conversation flow. Each state corresponds to different phases of the collaboration process:

\begin{itemize}[noitemsep,topsep=0pt,parsep=0pt,partopsep=0pt,leftmargin=1.5em]
    \item \textit{Discussion} ($s_d$): Agents engage in general dialogue, exchange ideas, and clarify task requirements.
    \item \textit{Synchronous task assignment} ($s_s$): Tasks are assigned to specific agents, pausing the group chat until completion (\cref{sec:framework-key-mech-task-assign}).
    \item \textit{Asynchronous task assignment} ($s_a$): Tasks are assigned without interrupting the ongoing discussion (\cref{sec:framework-key-mech-task-assign}).
    \item \textit{Pause \& trigger} ($s_p$): The group chat is paused, waiting for the completion of specified asynchronous tasks.
    \item \textit{Conclusion} ($s_c$): Marks the end of the collaboration, prompting a final summary.
\end{itemize}

These states align with speech acts in Speech Act Theory, such as assertives (discussion), directives (task assignment), commissives (pause \& trigger), and declarations (conclusion)~\citep{searle1976classification}.

\textbf{Autonomous State Transitions and Next Speaker Selection:} Recent studies have demonstrated the efficacy of LLMs in autonomously managing state transitions within predefined state spaces~\citep{DBLP:journals/corr/abs-2312-17445,DBLP:journals/corr/abs-2403-11322}, with state machines often enhancing overall system performance~\citep{DBLP:journals/corr/abs-2402-00798}. In \framework, the LLM within each client is tasked with determining state transitions and selecting the subsequent speaker. Let $\mathcal{M}_t$ be the set of messages exchanged up to time step $t$. We define the decision function of the LLM as:
$f_{\text{LLM}}: \mathcal{M}_t \times S \to S \times \mathcal{C}$,
where $S$ is the set of states and $\mathcal{C}$ is the set of clients. The next state $s_{t+1}$ and the next speaker $c_{t+1}$ are determined as:
$(s_{t+1}, c_{t+1}) = f_{\text{LLM}}(\mathcal{M}_t, s_t)$.
This decision-making process considers factors such as the completion of assigned tasks, the need for further discussion, and the overall goals of the collaboration. The autonomous selection of the next speaker ensures that the most relevant agents are involved at appropriate times, promoting efficient information exchange and problem-solving.

By implementing this autonomous conversation flow control mechanism, \framework enables structured and efficient communication among agents. This approach allows for dynamic adaptation to the needs of the collaboration, facilitating more effective problem-solving and decision-making in complex multi-agent scenarios.

\subsubsection{Task Assignment and Execution}
\label{sec:framework-key-mech-task-assign}

The task assignment and execution mechanism in \framework is designed to efficiently distribute work among agents and manage the execution of both simple and complex tasks. This mechanism works in concert with the team formation and conversation flow control mechanisms to ensure effective collaboration and task completion.

\textbf{Task Representation:} In \framework, a task $t \in \mathcal{T}$ is represented as a tuple $(d_t, \mathcal{S}_t)$, where $d_t$ is the task description and $\mathcal{S}_t = \{s_1, s_2, ..., s_n\}$ is the set of sub-tasks that $t$ can be decomposed into. Initially, $\mathcal{S}_t$ may be empty, with sub-tasks being identified dynamically during the collaboration process.

\textbf{Task Allocation:} Task allocation in \framework occurs within the context of group chats and is closely tied to the conversation flow control mechanism. There are two types of task allocation:
\begin{enumerate}[noitemsep,topsep=0pt,parsep=0pt,partopsep=0pt,leftmargin=1.5em]
    \item \textit{Synchronous Task Allocation:} When the group chat enters the synchronous task assignment state $s_s$, tasks are allocated to specific agents, and the group chat is paused until the tasks are completed. 

    \item \textit{Asynchronous Task Allocation:} In the asynchronous task assignment state $s_a$, tasks are allocated without interrupting the ongoing discussion. This allows for parallel execution of tasks.
\end{enumerate}
Formally, we can define a task allocation function $\alpha: \mathcal{T} \times \mathcal{G} \rightarrow \mathcal{P}(\mathcal{C})$, which maps a task and a group chat to a subset of clients responsible for executing the task.

\textbf{Task Execution:} Once a task is allocated, the responsible agent(s) begin execution. The execution process depends on the nature of the task and the capabilities of the agent. For integrated third-party agents, task execution is handled through the Agent Integration Block in the client's Foundation Layer. This block provides a standardized interface for task execution, typically in the form:
$\texttt{run}: \text{String} \rightarrow \text{TaskID}$,
where the input is the task description, and the output is a unique identifier for the task. Advanced features such as execution interruption could also be implemented in this stage. 

Upon completion of a task or sub-task, the responsible agent(s) report back to the group chat. In the case of synchronous tasks, this triggers the resumption of the group chat. For asynchronous tasks, the completion is noted, and any relevant information is shared with the group.

The pause \& trigger state $s_p$ in the conversation flow control mechanism plays a crucial role in managing the completion of multiple asynchronous tasks. It allows the group chat to wait for the completion of specified asynchronous tasks before proceeding, ensuring that all necessary information is available for subsequent stages of the collaboration.


\subsection{Comprehensive Message Protocol Design}
\label{sec:framework-protocol}

The effectiveness of the autonomous nested team formation and conversation flow control mechanisms in \framework relies on a comprehensive message protocol. This protocol enables seamless communication and collaboration among agents by encapsulating all necessary information required for various mechanisms to function properly.

\paragraph{Protocol Overview and Key Fields}
The agent message protocol in \framework is designed for extensibility and flexibility, facilitating effective multi-agent collaboration. The protocol consists of two main components: a header and a payload.

The header contains essential metadata about the message, ensuring correct addressing and processing by receiving agents. Key fields in the header include:
\begin{itemize}[noitemsep,topsep=0pt,parsep=0pt,partopsep=0pt,leftmargin=1.5em]
    \item \texttt{sender}: The unique identifier of the agent sending the message.
    \item \texttt{group\_id}: The identifier of the group chat to which the message belongs.
\end{itemize}

The payload carries the main content of the message, varying by message type. It can include:
\begin{itemize}[noitemsep,topsep=0pt,parsep=0pt,partopsep=0pt,leftmargin=1.5em]
    \item \texttt{message\_type}: Indicates the purpose of the message (e.g., discussion, task assignment, pause \& trigger).
    \item \texttt{next\_speaker}: The identifier(s) of the agent(s) expected to respond.
\end{itemize}

This structure contains other fields to support the diverse functionalities of \framework effectively. A detailed explanation and example of the message protocol can be found in~\cref{sec:appendix-ioa-detail-message-protocol}.

To ensure seamless communication and coordination, both the client and server components of \framework implement the message protocol. When a client sends a message, it encodes it according to the protocol and transmits it to the server. The server parses the message, extracts relevant information from the header, and routes it to the appropriate group chat based on the \texttt{group\_id}. Upon receiving a message, the client decodes it and processes it accordingly. This consistent implementation ensures that all agents can understand and respond to messages correctly, regardless of their roles or tasks, maintaining a coherent and efficient collaboration process.

\subsection{Putting It All Together: A Walkthrough of \framework in Action}
\label{sec:framework-mech-walkthrough}
\begin{figure}[t]
    \centering
    \includegraphics[width=\textwidth]{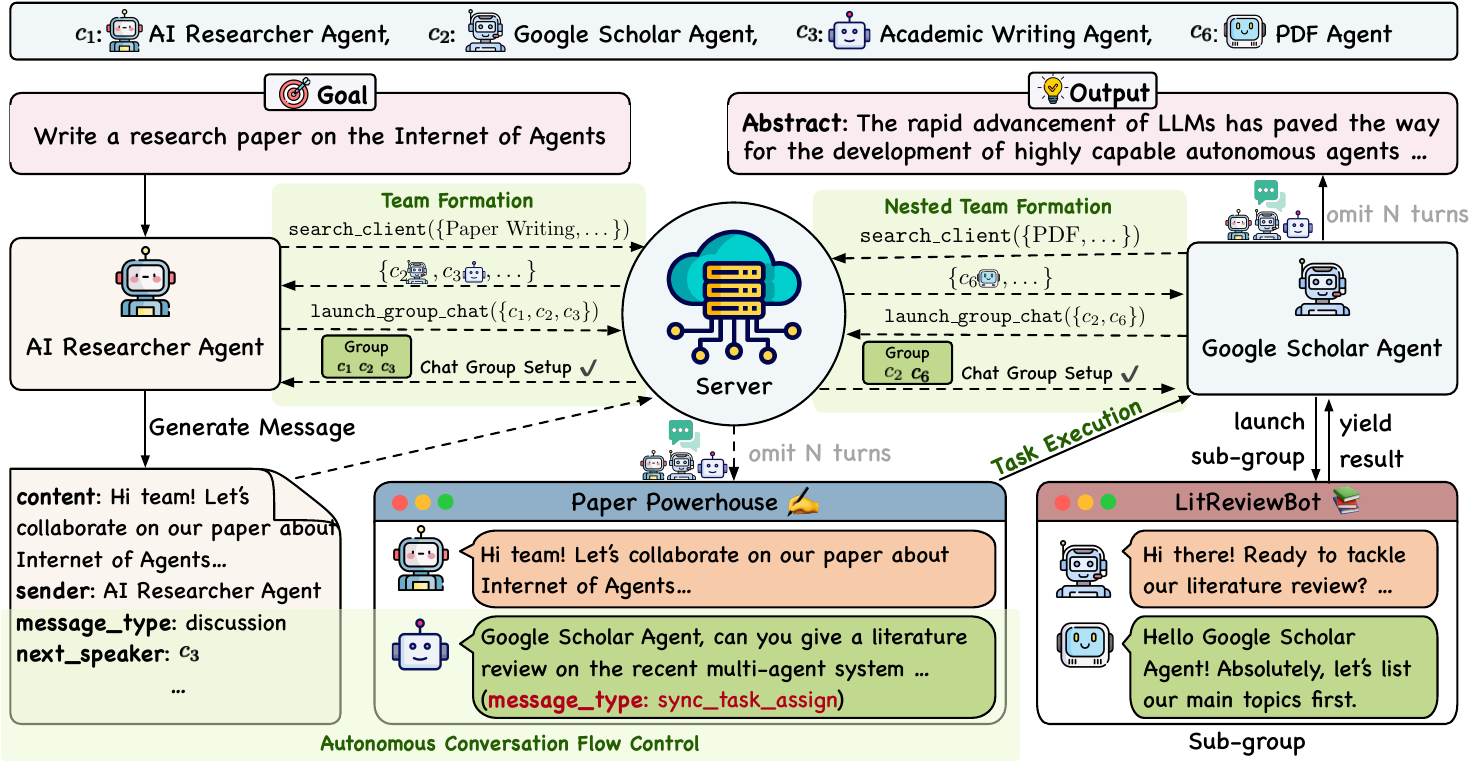}
    \caption{An example walkthrough of the major components of \framework.}
    \label{fig:walkthrough}
\end{figure}
To illustrate the integrated functionality of \framework, in~\cref{fig:walkthrough}, we present an example walkthrough of the system with an illustrative complex task: writing a research paper on the Internet of Agents. Initially, client $c_1$, an AI research specialist trained additionally on AI academic paper, engages the Team Formation Block, utilizing the $\texttt{search\_client}$ function with 
a list of keywords \{Internet, Multi-Agent System Specialist, Paper Writing, LLM Expert\}. The server returns a set of matched clients $\{c_2, c_3, c_4, c_5\}$, from which $c_1$ forms group $g_0$ with members $\{c_1, c_2, c_3\}$ via $\texttt{launch\_group\_chat}$, where $c_2$ has access to scholarly databases and $c_3$ specializes in academic writing.

Upon the formation of group chat $g_0$, all clients transition to the Communication Block for $g_0$, where the autonomous conversation flow control mechanism, implemented as a finite state machine, guides the collaboration. The process begins with brainstorming in the discussion state ($s_d$), progressing to task assignment states ($s_s, s_a$) where agents are allocated specific responsibilities. For instance, $c_2$ is tasked with conducting a literature review using its access to scholarly resources. The nested team formation mechanism is demonstrated when $c_2$ identifies a need for specialized PDF expertise. This prompts $c_2$ to initiate a sub-group formation process, resulting in the creation of sub-group $g_1$ with a new agent $c_6$, a PDF expert. Throughout the process, the conversation alternates between discussion ($s_d$) and asynchronous task assignment ($s_a$) states, facilitating parallel work on assigned tasks. The message protocol ensures efficient communication, enabling the exchange of ideas, citations, and draft segments across the nested group structure.

In the final integration phase, the group enters a synchronous task assignment state ($s_s$) for collaborative editing and refinement, demonstrating \framework's capacity for coordinating intensive, real-time collaboration among multiple agents. The process concludes with a transition to the conclusion state ($s_c$), where a final review is conducted and the paper is prepared for submission.

\section{Experiments}
\label{sec:experiments}
To demonstrate the effectiveness and versatility of \framework in integrating heterogeneous agents, we conduct comprehensive experiments across a diverse set of tasks. These experiments are designed to showcase different aspects of agent heterogeneity: tool variability (\cref{sec:experiments-general-ai-gaia}), architectural diversity (\cref{sec:experiments-general-ai-open-ended}), disparate observation and action spaces (\cref{sec:experiments-embodied}), and varied knowledge bases (\cref{sec:experiments-rag}). Our objective is twofold: first, to illustrate \framework's proficiency in facilitating collaboration among heterogeneous agents, and second, to highlight its adaptability across various problem domains. In this section, we present our experimental results and offer comparative analyses between \framework and state-of-the-art (SoTA) approaches for each task category. The prompts within \framework are kept the \textit{same} across different tasks, and are not specifically tuned for a certain task.\footnote{If not specified, we use GPT-4-1106-preview model in our experiments.}

\begin{table}[t]
    \centering
    \caption{The performance on the validation set of GAIA benchmark.}
    \resizebox{\textwidth}{!}{
    \begin{tabular}{l c c c c c}
        \toprule
        \textbf{Models} & \textbf{Agent Type} & \textbf{Level 1} & \textbf{Level 2} & \textbf{Level 3} & \textbf{Overall} \\
        \midrule
        GPT-4 & \faUser & 15.09 & 2.33 & 0.00 & 6.06\\
        GPT-4-Turbo & \faUser & 20.75 & 5.81 & 0.00 & 9.70\\
        \midrule
        AutoGPT-4~\citep{Significant_Gravitas_AutoGPT} & \faUser & 13.21 & 0.00 & 3.85 & 4.85\\
        GPT-4 + Plugins~\citep{DBLP:journals/corr/abs-2311-12983} & \faUser & 30.30 & 9.70 & 0.00 & 14.60\\
        FRIDAY~\citep{DBLP:journals/corr/abs-2402-07456} & \faUser & 45.28 & 34.88 & 11.54 & 34.55 \\
        \midrule
        AutoGen~\citep{DBLP:journals/corr/abs-2308-08155} & \faUsers & \textbf{54.72} & 38.37 & 11.54 & 39.39\\
        \framework & \faUsers & 50.94 & \textbf{40.70} & \textbf{15.38} & \textbf{40.00}\\
        \bottomrule
    \end{tabular}}
    \label{tab:gaia}
\end{table}

\subsection{Heterogeneous Tools: GAIA Benchmark}
\label{sec:experiments-general-ai-gaia}

To evaluate \framework's capability in integrating agents with heterogeneous tools, we employ the GAIA benchmark~\citep{DBLP:journals/corr/abs-2311-12983}. This benchmark comprises a diverse set of real-world questions designed to assess an agent system's proficiency in solving complex tasks through the synergistic application of multiple skills, including natural language understanding, reasoning, and external knowledge integration. The benchmark's three-tiered difficulty structure provides a robust testbed for evaluating the capability of agent systems.

\textbf{Experimental Setups:}
We instantiate \framework with four basic ReAct agents~\citep{DBLP:conf/iclr/YaoZYDSN023}, each equipped with a distinct tool: a web browser, a code interpreter, a Wikidata searcher, and a YouTube video transcript downloader. This configuration allows us to assess \framework's ability to orchestrate collaboration among agents with heterogeneous tools. We benchmark \framework against several SoTA agent systems, evaluating performance across all three difficulty levels of GAIA, as well as overall performance. Detailed implementation specifics are provided in~\cref{sec:appendix-exp-detail-gaia}.

\textbf{Results and Analysis:}
The experimental results, presented in~\cref{tab:gaia}, demonstrate \framework's superior performance across the GAIA benchmark. Despite utilizing only basic ReAct agents, \framework achieves the highest overall performance, surpassing all other approaches. Notably, \framework exhibits exceptional performance in the more challenging Level 2 and Level 3 tasks, which demand advanced reasoning and intricate collaboration. This performance underscores the efficacy of \framework's communication mechanisms and its capacity to facilitate seamless inter-agent collaboration.

In comparison to AutoGen, \framework demonstrates superior performance in two out of three difficulty levels. This superiority can be attributed to \framework's collaboration mechanisms and the flexibility of integrating agents with different tools, while in AutoGen, only one agent utilizes different tools, and other agents act as feedback providers. The mechanisms implemented in \framework enable adaptive team composition and efficient sub-task execution, culminating in enhanced performance on complex, multi-faceted problems.

The results from the GAIA benchmark underscore \framework's potential as a powerful orchestrator for diverse agents in solving real-world, multi-step problems. By providing a flexible and efficient platform for agent collaboration, \framework enables even basic agents to achieve SoTA performance, outperforming more sophisticated standalone agents. This outcome highlights the critical role of effective communication and coordination in multi-agent systems and validates the architectural and design choices underpinning \framework.

\subsection{Heterogeneous Architecture: Open-Ended Instruction Benchmark}
\label{sec:experiments-general-ai-open-ended}
\begin{figure}[t]
    \centering
    \includegraphics[width=0.9\textwidth]{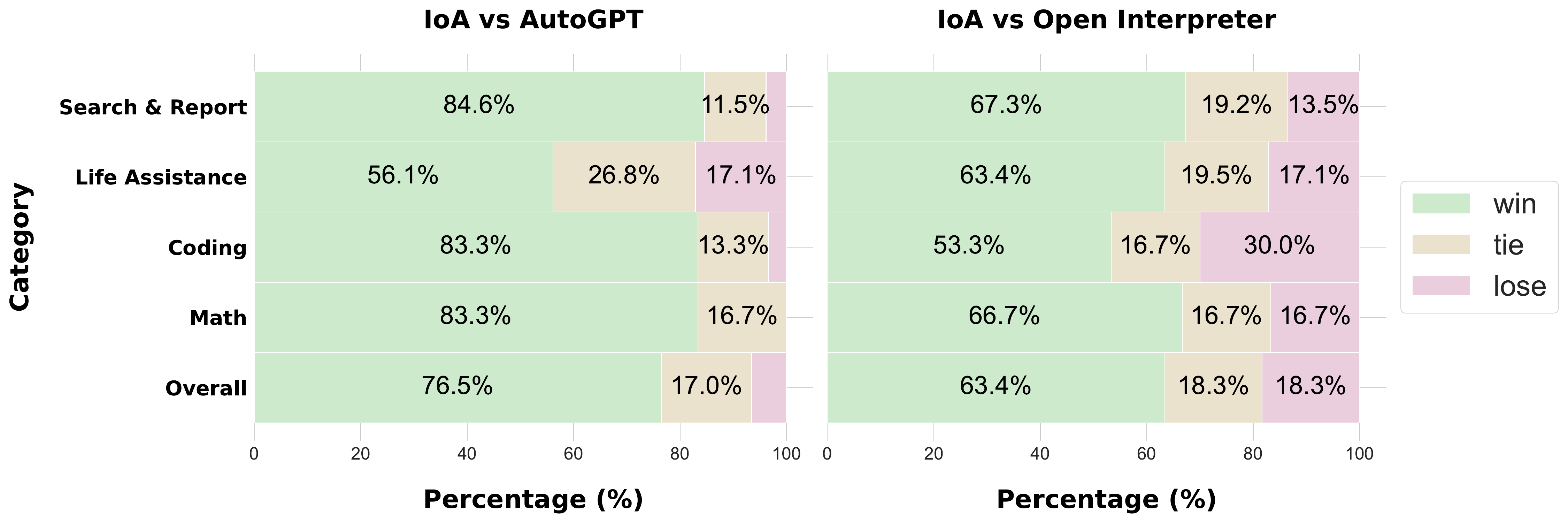}
    \caption{Comparison of win rates on the open-ended instruction benchmark between \framework, AutoGPT, and Open Interpreter.}
    \label{fig:open-ended}
\end{figure}

To evaluate \framework's capability in integrating and orchestrating agents with heterogeneous architectures, we develop a comprehensive benchmark comprising 153 open-ended instructions with self-instruct~\citep{DBLP:conf/acl/WangKMLSKH23}. This benchmark spans four diverse categories: search \& report, coding, mathematics, and life assistance. Unlike the GAIA benchmark, which primarily focuses on question-answering tasks with deterministic answers, our curated benchmark incorporates a higher proportion of non-QA tasks requiring generative responses. This design choice aims to better reflect the diverse nature of real-world challenges that agent systems are expected to address. The curation process is elaborated at~\cref{sec:appendix-exp-detail-open-ended}.

\textbf{Experimental Setups:} 
In this experimental setup, we integrate two SoTA third-party agents with distinct architectures: AutoGPT~\citep{Significant_Gravitas_AutoGPT} and Open Interpreter~\citep{OpenInterpreter}, into the \framework ecosystem. The integration process, detailed in~\cref{sec:appendix-exp-detail-open-ended}, demonstrates \framework's versatility in accommodating agents with divergent internal structures and operational paradigms. This configuration allows us to assess \framework's efficacy in facilitating collaboration among independently developed agents with heterogeneous architectures.

For evaluation, we employ GPT-4-1106-preview as an impartial judge, a choice supported by previous research demonstrating high agreement between GPT models and human evaluators in assessing response quality~\citep{vicuna2023,DBLP:conf/nips/ZhengC00WZL0LXZ23,DBLP:journals/corr/abs-2308-07201}. To mitigate potential order-induced biases, we implement a robust evaluation approach following~\citet{DBLP:conf/nips/ZhengC00WZL0LXZ23}, where the order of responses is alternated in the prompt. A "win" is only declared when one competitor is consistently judged superior across both orderings.

\textbf{Results and Analysis:} 
The experimental results, illustrated in~\cref{fig:open-ended}, demonstrate \framework's significant performance advantages when orchestrating the collaboration between AutoGPT and Open Interpreter. \framework consistently outperforms both individual agents across all four task categories. Overall, \framework achieves a remarkable win rate of 76.5\% against AutoGPT and 63.4\% against Open Interpreter. These results underscore \framework's proficiency in efficiently gathering and synthesizing information, as well as its effectiveness in facilitating collaborative problem-solving across diverse domains. 

The demonstrated capability of \framework to seamlessly integrate and orchestrate agents with heterogeneous architectures enables the harness of the strengths of diverse, independently developed agents, making it possible to create more versatile and capable agent systems. As the landscape of specialized AI agents continues to expand, \framework's potential to integrate and facilitate collaboration among these diverse entities positions it as a promising platform for the development of increasingly sophisticated and adaptive agent systems.

\subsection{Heterogeneous Observation and Action Space: Embodied Agent Tasks}
\label{sec:experiments-embodied}
\begin{table}[t]
    \centering
    \caption{Average success rate and the number of steps on different tasks from RoCoBench.}
    \begin{tabular}{c c c c c c c}
    \toprule
         \textbf{Model} & \textbf{Metric} & \textbf{Cabinet} & \textbf{Sweep} & \textbf{Sandwich} & \textbf{Sort} & \textbf{Rope} \\
         \midrule
         \multirow{2}{*}{\shortstack{Central Plan\\(oracle)}} & {\small Success} & 0.90 & \textbf{1.00} & 0.96 & 0.70 & 0.50\\
         & {\small \#Step} & \underline{4.0} & 8.4 & \underline{8.8} & 8.6 & \underline{2.3}\\
         \midrule
         \multirow{2}{*}{\shortstack{Roco\\Dialog}} & {\small Success} & 0.75 & 0.70 & 0.70 & 0.70 & \textbf{0.70} \\
         & {\small \#Step} & 4.7 & \underline{7.9} & 9.1 & \underline{5.4} & 2.4\\
         \midrule
         \multirow{2}{*}{\framework} & {\small Success} & \textbf{1.00} & 0.80 & \textbf{1.00} & \textbf{1.00} & \textbf{0.70}\\
         & {\small \#Step} & 4.6 & 8.5 & 8.9 & 5.8 & 2.6\\
         \bottomrule
    \end{tabular}
    \label{tab:roco}
\end{table}

To evaluate \framework's efficacy in orchestrating agents with heterogeneous observation and action spaces, we conduct experiments in the domain of embodied AI. This domain presents unique challenges, requiring agents to perceive, understand, and interact with their physical environment. We utilize RoCoBench~\citep{DBLP:journals/corr/abs-2307-04738}, a state-of-the-art benchmark designed to assess the collaboration and communication capabilities of embodied agents. RoCoBench comprises six collaborative tasks, each mandating two or three agents with partial, often distinct action space or observations of the environment to cooperate towards a common objective.

\textbf{Experimental Setups:} 
We benchmark \framework against two baselines established by \citet{DBLP:journals/corr/abs-2307-04738}: (1) Central Plan, a centralized agent has complete environmental information and control over all embodied agents, and (2) Roco Dialog, a specialized multi-agent framework designed for this task, enabling agent communication and decision-making.

Given that RoCoBench requires agents to output action plans in a specific format rather than interact with tools, we adapt \framework to this scenario without integrating external agents. Instead, we provide environmental observations to two \framework clients and extract their action plans from their discussion. This setup allows us to evaluate \framework's ability to manage agents with heterogeneous observation and action spaces. Detailed implementation specifics are available in~\cref{sec:appendix-exp-detail-embodied}.
To ensure a fair comparison, we conduct 10 runs for both \framework and Roco Dialog for each task, reporting average success rates and steps taken. Results for Central Plan are sourced directly from \citet{DBLP:journals/corr/abs-2307-04738}. Note that the Pack Grocery task is omitted due to implementation errors in the benchmark release.

\textbf{Results and Analysis:} 
\cref{tab:roco} presents the average success rates and steps required for task completion. Remarkably, despite not being specifically optimized for embodied tasks, \framework outperforms Roco Dialog, a framework tailored for this benchmark, in four out of five tasks in terms of success rate. \framework achieves perfect scores on the Cabinet, Sandwich, and Sort tasks, demonstrating the robustness of its communication and collaboration mechanisms in enabling embodied agents with heterogeneous observation and action spaces to work synergistically towards common goals. Even more impressive is \framework's performance relative to the Central Plan baseline, which benefits from full environmental observability. \framework's success rates are superior or comparable to Central Plan across tasks, although it generally requires slightly more decision steps for task completion. Given that \framework is a general multi-agent framework not specifically designed for embodied AI tasks, the marginal increase in step count is a reasonable trade-off for its versatility and effectiveness.


The success of \framework in this embodied AI scenario highlights its versatility and effectiveness. It suggests that the principles underlying \framework, e.g., autonomous conversation flow control, are fundamentally generalizable, indicating \framework's potential applicability in a wide range of real-world scenarios where agents must collaborate despite having different perspectives or capabilities.

\subsection{Heterogeneous Knowledge: Retrieval-Augmented Generation}
\label{sec:experiments-rag}

To evaluate \framework's efficacy in orchestrating agents with heterogeneous knowledge, we conduct experiments on retrieval-augmented generation (RAG) tasks~\citep{lewis2021retrievalaugmented}. RAG tasks present a unique challenge where agents must retrieve relevant information from diverse sources and collaborate to synthesize accurate responses, making them an ideal testbed for assessing \framework's ability to manage knowledge heterogeneity and facilitate effective inter-agent communication.

\textbf{Experimental Setups:} 
We implement \framework with GPT-3.5-turbo-0125 as the core language model, following Apollo's Oracle~\citep{Wang2023ApollosOR}. To evaluate knowledge heterogeneity and its impact, we design three scenarios: 1) \textit{Heterogeneous Knowledge}: Two clients access different evidence pools (Wikipedia/Google), testing \framework's ability to manage knowledge heterogeneity. 2) \textit{Homogeneous Knowledge (2 Agents)}: Two clients access both pools, serving as a control to isolate heterogeneity effects. 3) \textit{Homogeneous Knowledge (3 Agents)}: Three clients access both pools, assessing scalability and knowledge redundancy trade-offs.

This design allows us to disentangle the effects of knowledge heterogeneity from agent count and knowledge redundancy. We evaluate across four datasets: TriviaQA~\citep{joshi2017triviaqa}, Natural Questions (NQ)~\citep{Kwiatkowski2019NaturalQA}, HotpotQA~\citep{yang2018hotpotqa}, and 2WikiMultiHopQA (2WMHQA)~\citep{ho2020constructing}, using 250 randomly sampled question-answer pairs from each. Implementation details are in \cref{sec:appendix-exp-detail-rag}.

\begin{table*}[t]
    \centering
    \resizebox{\linewidth}{!}{
    \begin{tabular}{l>{}ccccccc}
        \toprule
        \textbf{Model} & \textbf{TriviaQA} & \textbf{NQ} & \textbf{HotpotQA} & \textbf{2WMHQA} & \textbf{Overall} \\
        \midrule
        GPT 4 & 0.902 & 0.692 & 0.566 & 0.284 & 0.611 \\
        \midrule
        GPT 3.5 Turbo & 0.778 & 0.532 & 0.384 & 0.210 & 0.476  \\
        \quad + Zero-Shot CoT~\citep{DBLP:conf/nips/Wei0SBIXCLZ22} & 0.772 & 0.588 & 0.410 & 0.190 & 0.490 \\
        \quad + Self Consistency~\citep{DBLP:conf/iclr/0002WSLCNCZ23} & 0.818 & 0.622 & 0.408 & 0.206 & 0.514 \\
        \quad + Reflxion~\citep{DBLP:conf/nips/ShinnCGNY23} & 0.762 & 0.586 & 0.378 & 0.254 & 0.495 \\
        \quad + Multi-Agent  Debate1~\citep{DBLP:journals/corr/abs-2305-14325} & 0.798 & 0.648 & 0.394 & 0.186 & 0.507 \\
        \quad + Multi-Agent  Debate2~\citep{DBLP:journals/corr/abs-2305-19118} & 0.756 & 0.576 & 0.450 & 0.334 & 0.529 \\
        \midrule
        Apollo's Oracle (Homogeneous) & \underline{0.834} & 0.662 & 0.542 & 0.350 & 0.597 \\
        \midrule
        \framework + 2 Agents (Heterogeneous) & 0.803 & \textbf{0.708} & 0.478 & 0.449 & 0.610 \\
        \framework + 2 Agents (Homogeneous) & 0.820 & 0.671 & \textbf{0.586} & \textbf{0.530} & \underline{0.652} \\ 
        \framework + 3 Agents (Homogeneous)& \textbf{0.908} & \underline{0.682} & \underline{0.575} & \underline{0.519} & \textbf{0.671} \\
        \bottomrule
    \end{tabular}
    }
    \caption{Results for RAG task. \framework, based on GPT-3.5, performs on par with or better than GPT-4 across all tasks. Best results (excluding GPT-4) are in bold, and second-best are underlined. \textit{Heterogeneous} means agents have different evidence pools, while \textit{Homogeneous} means all agents access all evidence pools.}
    \label{tab:rag}
\end{table*}

\textbf{Results and Analysis:} 
\cref{tab:rag} demonstrates \framework's remarkable performance across all datasets, often surpassing or matching GPT-4 despite being based on GPT-3.5. On two out of four tasks, \framework's heterogeneous knowledge scenario outperforms homogeneous Apollo's Oracle, showcasing \framework's effectiveness in managing knowledge diversity. This configuration achieves the best performance on NQ and competitive results on other datasets, often outperforming single-model approaches and specialized frameworks like Apollo's Oracle. This underscores \framework's efficacy in facilitating information exchange and synthesis from heterogeneous sources, effectively compensating for individual agents' knowledge gaps.

We also conduct experiments in homogeneous settings. \framework with 3 agents achieves the best overall performance, outperforming all baselines on TriviaQA and showing competitive results on other datasets. Interestingly, the 2-agent homogeneous configuration outperforms the 3-agent setup on HotpotQA and 2WikiMultiHopQA, suggesting that optimal agent configuration may be task-dependent. These results not only validate \framework's effectiveness in RAG tasks but also highlight its potential as a versatile platform for managing both heterogeneous and homogeneous knowledge in multi-agent systems.
\section{Analysis}
\label{sec:analysis}

\subsection{Team Formation Precision}
\label{sec:analysis-teamup}

To evaluate the precision of \framework's autonomous team formation mechanism, we developed a benchmark using GPT-4, comprising 625 diverse tasks paired with 1500 dummy agent profiles. This simulated environment allows us to assess the accuracy of both regular and nested team formation in a large-scale setting. Detailed data construction processes are available in~\cref{sec:appendix-teamup-detail}.

\paragraph{Experimental Design:} 
We evaluate two distinct scenarios: regular team formation and nested team formation. For regular team formation, each task is associated with 2 or more suitable agent profiles generated by GPT. For nested team formation, we generate a subtask for each original task that can or cannot be completed by the initially formed team, if not, an additional agent profile capable of addressing this subtask is generated. We evaluate whether the team can correctly decide when to enter the nested team formation stage, and evaluate the precision of the nested team formation.

We assess both settings using four metrics: Top@1 and Top@10 recall rates, Mean Rank (MR), and Mean Reciprocal Rank (MRR). Top@1 measures exact matches, while Top@10 accounts for semantic similarity, considering an agent as recalled if a recruited agent is among the top 10 most similar to a labeled agent. MR and MRR provide insights into the ranking quality of retrieved agents.

\begin{wraptable}{r}{0.55\textwidth}
    \centering
    \vspace{-1.2em}
    \caption{Performance of Team Formation Mechanisms. \textit{Regular} denotes the initial team formation setting, and \textit{Nested} denotes the nested team formation mechanism.}
    \begin{tabular}{c c c c c}
    \toprule
         \textbf{} & \textbf{Top@1$\uparrow$} & \textbf{Top@10$\uparrow$} & \textbf{MR$\downarrow$} & \textbf{MRR$\uparrow$}\\
         \midrule
         \multirow{1}{*}{Regular} & 41.4\% & 64.9\% & 27.4 & 50.1\%\\
         \multirow{1}{*}{Nested} & 59.7\% & 81.8\% & 10.6 & 66.5\% \\
         \bottomrule
    \end{tabular}
    \vspace{-1em}
    \label{tab:teamup}
\end{wraptable}

\textbf{Results and Analysis:}
Table~\ref{tab:teamup} presents the performance of both team formation mechanisms, each evaluated on its own specific dataset and setting. In the regular team formation scenario, which assesses the ability to form initial teams for given tasks, we observe a Top@1 recall of 41.4\% and a Top@10 recall of 64.9\%. This indicates that the mechanism can exactly match the labeled agents 41.4\% of the time, and when considering semantic similarity, the retrieved agent fall into the top 10 similar agents to the labeled agent for 64.9\% of the time. The Mean Rank (MR) of 27.4 and Mean Reciprocal Rank (MRR) of 50.1\% suggest that, on average, relevant agents are ranked within the top 30 results, with a tendency towards high ranking.

For the nested team formation scenario, which evaluates the mechanism's performance in a setting where subtasks may emerge requiring additional expertise, we see a Top@1 recall of 59.7\% and a Top@10 recall of 81.8\%. The MR of 10.6 and MRR of 66.5\% indicate that relevant agents are typically found within the top 11 results, with a strong tendency towards very high rankings. These metrics suggest effective performance in this more dynamic setting.

These results demonstrate \framework's capability to form precise teams in both initial task allocation and in scenarios where task requirements may evolve. The high recall rates, especially with similarity matching (Top@10), are crucial for addressing complex tasks that require diverse or specialized skills.

\subsection{Cost and Sub-Optimal Communication Pattern Analysis}
\label{sec:analysis-cost}

To evaluate the economic feasibility and potential for optimization of the \framework, we conduct a cost analysis on the open-ended instruction benchmark (\cref{sec:experiments-general-ai-open-ended}), where AutoGPT and Open Interpreter are integrated. We compare the average cost per task for these agents when operating individually and when integrated into the \framework.

\begin{wraptable}{r}{0.5\textwidth}
    \centering
    \vspace{-1.3em}
    \setlength{\tabcolsep}{3pt}
    \caption{Cost analysis of standalone agents and \framework-integrated agents on the open-ended instruction benchmark.}
    \begin{tabular}{l c c c c c c}
        \toprule
        \textbf{Setting} & \textbf{Cost per Task} \\
        \midrule
        AutoGPT (Standalone) & \$0.39 \\
        Open Interpreter (Standalone) & \$0.16 \\
        \midrule
        AutoGPT (in \framework) & \$0.33 \\
        Open Interpreter (in \framework) & \$0.13 \\
        \framework Communication & \$0.53 \\
        \framework Communication (Dedup.) & \$0.28 \\
        \midrule
        \framework Overall & \$0.99 \\
        \framework Overall (Dedup.) & \$0.74 \\
        \bottomrule
    \end{tabular}
    \vspace{-1.5em}
    \label{tab:cost-analysis}
\end{wraptable}

As shown in~\cref{tab:cost-analysis}, when integrated into \framework, the costs of both agents are decreased due to the task decomposition for each task. However, the \framework introduces an additional communication cost of \$0.53 per task, resulting in an overall cost of \$0.99.

\begin{wrapfigure}{r}{0.4\textwidth}
    \centering
    \includegraphics[width=0.4\textwidth]{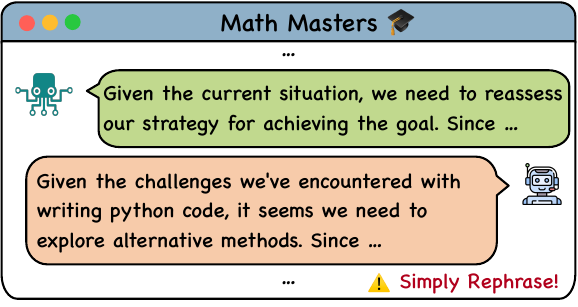}
    \caption{An example of the repeated communication.}
    \vspace{-1em}
    \label{fig:suboptimal}
\end{wrapfigure}
During our analysis, we observed unexpected and suboptimal communication patterns that contributed to the high communication cost. One notable pattern was the repetition of information, where the LLMs in the clients would repeat or rephrase previous chats from themselves or others, leading to a stagnation in progress. This phenomenon was particularly prevalent after several asynchronous task assignments. Although each task assignment did not require immediate waiting, as the conversation progressed, new decisions had to be made based on the conclusions from previously assigned and not yet completed asynchronous tasks. Despite providing the client LLMs with the option to switch the group chat state to pause \& trigger, they sometimes fail to switch, as illustrated in~\cref{fig:suboptimal}. This drawback in LLM is also observed in other multi-agent work~\citep{DBLP:conf/nips/LiHIKG23,DBLP:journals/corr/abs-2307-04738}.

To quantify the impact of this suboptimal communication pattern, we manually removed the repetitions and recalculated the token numbers and corresponding costs. Surprisingly, this resulted in a nearly 50\% reduction in communication costs, as shown in the "Dedup." rows of~\cref{tab:cost-analysis}. This finding aligns with observations from other multi-agent communication frameworks, suggesting that while modern LLMs are well-aligned to be effective chatbot assistants, they may not be optimally aligned to be efficient communicating agents. Agents should not only complete the given tasks accurately but also communicate effectively with others, understanding conversation states and making proper decisions. This insight raises new research questions regarding the agent alignment of LLMs and highlights the need for further investigation in this area.

Despite the current cost overhead and suboptimal communication patterns, the \framework demonstrates significant potential for enabling effective collaboration among heterogeneous agents. By addressing these challenges through prompt optimization, protocol refinement, and the development of more sophisticated frameworks under the concept of \framework, we believe that the cost of communication can be significantly reduced. As research progresses, \framework and similar approaches will become increasingly attractive and economically viable solutions for complex multi-agent systems.

\section{Related Work}
\label{sec:related-work}
\paragraph{LLM-based Agents}
Recent advancements in LLMs, such as GPT~\citep{DBLP:journals/corr/abs-2303-08774}, Claude~\citep{anthropic2024claude} and Gemini~\citep{DBLP:journals/corr/abs-2403-05530}, have led to the development of highly capable AI agents, which can engage in natural language interactions and perform a wide range of tasks. To enhance the capabilities of LLM-based agents, researchers have explored the integration of external tools and knowledge sources~\citep{DBLP:journals/corr/abs-2112-09332,DBLP:conf/iclr/YaoZYDSN023,DBLP:conf/nips/SchickDDRLHZCS23,DBLP:conf/nips/0001ST00Z23}, enabling agents to access and utilize relevant information beyond their pre-trained knowledge. The various agents have demonstrated significant progress in a wide range of domains, including operating system interactions, software engineering, and general AI applications. For instance, OS-Copilot facilitates generalist interactions across web browsers and code terminals~\citep{DBLP:journals/corr/abs-2402-07456}, while OpenDevin focuses on autonomous software development tasks such as coding and debugging~\citep{opendevin2024}. Other notable developments include XAgent for complex task solving~\citep{xagent2023} and Voyager~\citep{DBLP:journals/corr/abs-2305-16291}, an open-ended embodied agent leveraging LLMs for Minecraft game-playing. These advancements have laid the foundation for more sophisticated and versatile LLM-based agents, capable of autonomous task execution and continuous learning.

\paragraph{LLM-based Multi-Agent Systems}
Building upon the success of individual LLM-based agents, researchers have begun to explore the potential of multi-agent systems composed of these agents. Early works demonstrated the feasibility of using LLMs to simulate multi-agent interactions and emergent behaviors~\citep{DBLP:conf/uist/ParkOCMLB23}. Since then, various approaches have been proposed to enable effective collaboration and communication among LLM-based agents. Frameworks such as AgentVerse~\citep{chen2023agentverse} and AutoGen~\citep{DBLP:journals/corr/abs-2308-08155} provide the necessary infrastructure for agent collaboration. In software development, multi-agent systems like ChatDev~\citep{DBLP:journals/corr/abs-2307-07924}, MetaGPT~\citep{DBLP:journals/corr/abs-2308-00352} have shown promising results in automating coding, testing, and debugging processes. Despite these advancements, significant limitations remain, such as the lack of support for integrating diverse third-party agents, the inability to support distributed multi-agent systems, and the reliance on hard-coded communication protocols and state transitions. \framework aims to address these limitations and provide a more flexible and scalable platform for LLM-based multi-agent collaboration, paving the way for more advanced and practical systems that can tackle complex real-world problems effectively.
\section{Conclusion}
\label{sec:conclusion}
In this paper, we introduced \framework, a novel framework for LLM-based multi-agent collaboration inspired by the concept of the Internet. \framework addresses the limitations of existing multi-agent frameworks by providing a flexible and scalable platform for integrating diverse third-party agents, enabling distributed multi-agent collaboration, and introducing dynamic mechanisms for agent teaming and conversation flow control. Through extensive experiments on various benchmarks, we demonstrated the effectiveness of \framework in facilitating efficient collaboration among heterogeneous agents, consistently outperforming state-of-the-art baselines. 
As the field of LLM-based agents continues to advance, we believe that \framework will serve as a foundation for future research and development in multi-agent collaboration. By enabling the integration of diverse agents with specialized skills and knowledge, our framework opens up new possibilities for leveraging existing agents that were developed independently. We hope that our work will inspire further research in this promising direction and contribute to the development of more advanced and impactful multi-agent systems.

\bibliography{iclr2024_conference}
\bibliographystyle{iclr2024_conference}

\appendix

\section{Implementation Details of \framework}
\label{sec:appendix-ioa-detail}

In this appendix, we provide a comprehensive overview of the implementation details for each module in the client and server layers of \framework.

\subsection{Message Protocol}
\label{sec:appendix-ioa-detail-message-protocol}

\begin{wrapfigure}{r}{0.4\textwidth}
    \centering
    \vspace{-1.2em}
    \includegraphics[width=0.4\textwidth]{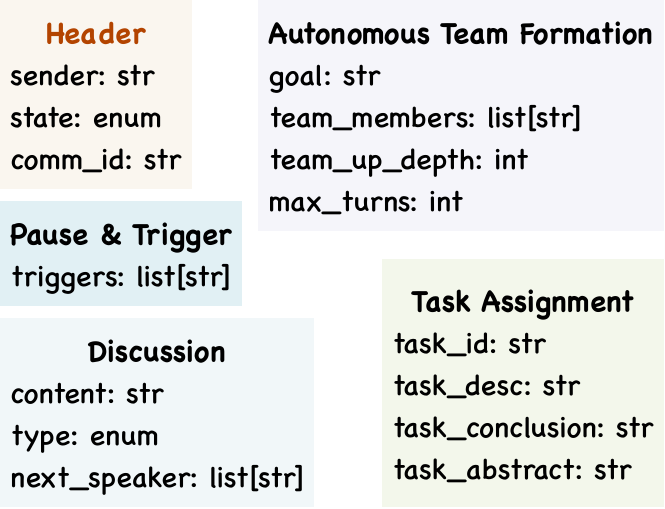}
    \caption{Fields in the \framework message protocol.}
    \vspace{-1em}
    \label{fig:protocol}
\end{wrapfigure}
To support the functionalities of \framework introduced in \cref{sec:framework-protocol}, we have designed a comprehensive agent message protocol that facilitates efficient communication and coordination among agents. The protocol, as illustrated in \cref{fig:protocol}, consists of several fields that cater to the specific requirements of various mechanisms within the framework.

Firstly, the protocol includes the following header for all message types:
\begin{itemize}[noitemsep,topsep=0pt,parsep=0pt,partopsep=0pt,leftmargin=1.5em]
    \item \texttt{sender} (str): The name or unique identifier of the agent sending the message.
    \item \texttt{state} (enum): The current state of the group chat associated with the message, which can be either team formation or communication.
    \item \texttt{comm\_id} (str): The unique identifier of the group chat to which the message belongs.
\end{itemize}

To support the autonomous team formation mechanism, the protocol incorporates the following fields:
\begin{itemize}[noitemsep,topsep=0pt,parsep=0pt,partopsep=0pt,leftmargin=1.5em]
    \item \texttt{goal} (str): The objective or task that the current group chat aims to accomplish.
    \item \texttt{team\_members} (list[str]): The names or unique identifiers of the agents required for the current group chat.
    \item \texttt{team\_up\_depth} (int): The depth of the current nested team formation, used to determine if the maximum allowed depth has been reached.
    \item \texttt{max\_turns} (int): The maximum number of discussion turns allowed for the current group chat. If exceeded, the group chat will be forced into the conclusion phase.
\end{itemize}

For facilitating the discussion phase, the protocol includes the following fields:
\begin{itemize}[noitemsep,topsep=0pt,parsep=0pt,partopsep=0pt,leftmargin=1.5em]
    \item \texttt{content} (str): The actual content of the current message.
    \item \texttt{type} (enum): Specifies the next dialogue state, which can be discussion, task assignment, or conclusion.
    \item \texttt{next\_speaker} (list[str]): The name(s) or unique identifier(s) of the agent(s) expected to speak next. In the discussion state, \texttt{next\_speaker} is limited to a single agent, while in the task assignment state, it can include multiple agents, indicating that the current message contains multiple task assignments.
\end{itemize}

To support the task assignment mechanism, the protocol incorporates the following fields:
\begin{itemize}[noitemsep,topsep=0pt,parsep=0pt,partopsep=0pt,leftmargin=1.5em]
    \item \texttt{task\_id} (str): The automatically generated unique identifier for the current task.
    \item \texttt{task\_desc} (str): The description of the task assigned to the client, extracted from the chat.
    \item \texttt{task\_conclusion} (str): The conclusion or result provided by the client after completing the assigned task.
    \item \texttt{task\_abstract} (str): A concise summary of the completed task.
\end{itemize}

Lastly, to support the pause \& trigger mechanism, the protocol includes the following field:
\begin{itemize}[noitemsep,topsep=0pt,parsep=0pt,partopsep=0pt,leftmargin=2em]
    \item \texttt{triggers} (list[str]): A list of task IDs that require a trigger to be set.
\end{itemize}

By adhering to this comprehensive agent message protocol for sending and receiving messages, clients within \framework can effectively achieve autonomous team formation and conversation flow control. The protocol ensures that all necessary information is communicated among agents, enabling seamless collaboration and coordination in various task scenarios.

\subsection{Client}
\label{sec:appendix-ioa-detail-client}

The client component of \framework plays a crucial role in enabling the integration and collaboration of heterogeneous agents. It consists of three layers: the Foundation Layer, the Data Layer, and the Interaction Layer. Each layer comprises several modules that work together to facilitate efficient communication, data management, and agent coordination. In this subsection, we provide a detailed overview of the implementation of each module within the client's layers.

\subsubsection{Foundation Layer}
\label{sec:appendix-ioa-detail-client-foundation}

\paragraph{Network Infrastructure Module}
In \framework, all clients maintain a persistent connection to the server using the WebSocket protocol, similar to an instant messaging application. When a client sends a message, it is transmitted to the server, which parses the \texttt{comm\_id} field in the message and forwards it to the other clients in the corresponding group chat via their respective WebSocket connections. The real-time nature of WebSocket ensures that messages are delivered promptly, enabling clients to receive and respond to messages without delay.

\paragraph{Data Infrastructure Module}
To support the data storage and retrieval requirements of the upper-level Data Layer modules, we employ SQLite as the primary database solution. SQLite provides a lightweight and efficient means of persisting and accessing data related to agent contacts, group information, and task management. By leveraging SQLite, the client can store and retrieve information about encountered agents, group chat details, and task assignments, ensuring data consistency and availability throughout the collaboration process.

\paragraph{Agent Integration Module}
The Agent Integration Module defines the protocol that third-party agents must adhere to in order to seamlessly integrate with \framework. Currently, the agent integration protocol in \framework requires agents to implement a function \texttt{def run(task\_desc: str) -> str}, which accepts a task description as input and returns a summary of the task completion. This simple yet effective protocol allows diverse agents to be incorporated into the framework, enabling them to contribute their unique capabilities to the collaboration process. As \framework evolves, the integration protocol can be extended to support more advanced functionalities and interaction patterns.

\subsubsection{Data Layer}
\label{sec:appendix-ioa-detail-client-data}

\paragraph{Agent Contact Module}
The Agent Contact Module is responsible for maintaining a record of the clients that the current client has previously collaborated with. It stores information such as the names and descriptions of these clients, providing a valuable reference for future collaborations. The module aims to support the client in evaluating and storing collaboration outcomes after each task, allowing it to make informed decisions when forming teams for subsequent tasks. During the team formation process, the information stored in this module is included in the prompt to assist the client in selecting the most suitable partners based on prior experiences.

\paragraph{Group Info Module}
The Group Info Module manages all group chat-related information, including the following fields:

\begin{itemize}
    \item \texttt{comm\_id} (str): The unique identifier of the group chat.
    \item \texttt{goal} (str): The objective or task that the group chat aims to accomplish.
    \item \texttt{team\_members} (str): The list of agents participating in the group chat.
    \item \texttt{state} (str): The current state of the group chat (e.g., team formation, discussion, task assignment, conclusion).
    \item \texttt{conclusion} (str | None): The final outcome or conclusion reached by the group chat.
    \item \texttt{team\_up\_depth} (int): The depth of the nested team formation within the group chat.
    \item \texttt{max\_turns} (int): The maximum number of communication turns allowed in the group chat.
\end{itemize}

By organizing and persisting this information, the Group Info Module enables clients to maintain a coherent view of the ongoing collaborations and their progress.

\paragraph{Task Management Module}
The Task Management Module is responsible for storing and tracking the tasks assigned within each group chat. It maintains the following fields for each task:

\begin{itemize}
    \item \texttt{task\_id} (str): The unique identifier of the task.
    \item \texttt{task\_desc} (str): The detailed description of the task.
    \item \texttt{task\_abstract} (str): A concise summary of the task.
    \item \texttt{assignee} (str): The agent assigned to complete the task.
    \item \texttt{status} (enum): The current status of the task (e.g., pending, in progress, completed).
    \item \texttt{conclusion} (str | None): The final result or outcome of the task.
\end{itemize}

By keeping track of task-related information, the Task Management Module enables clients to monitor the progress of assigned tasks and ensures that all task-related data is readily available for reference and decision-making purposes.

\subsubsection{Interaction Layer}
\label{sec:appendix-ioa-detail-client-interaction}

\paragraph{Team Formation Module}
As briefly introduced in \cref{sec:framework-key-mech-teamup}, when a client receives a task, it is equipped with two essential tools: \texttt{search\_agent(desc: list[str]) -> list[agent]} and \texttt{launch\_group\_chat(team\_members: list[str] | None) -> comm\_id}. The client must decide whether to utilize the \texttt{search\_agent} tool to find agents on the server that match the specified description, or to directly call the \texttt{launch\_group\_chat} tool based on the discovered agents and historical collaboration information. If the client invokes \texttt{launch\_group\_chat} without specifying any agents, it implies that the task will be completed by a single agent. To prevent infinite loops, \framework imposes a limit on the maximum number of tool calls, set to 10 by default. If the client reaches this limit without successfully launching a group chat, it is forced to invoke the \texttt{launch\_group\_chat} tool to initiate the collaboration process.

\paragraph{Communication Module}
The Communication Module handles the core functionalities of message generation and message reception. When a client generates a message, \framework processes it according to the agent message protocol. If the message type is \texttt{conclusion}, the client enters the conclusion phase, where it provides a final answer to the group chat goal based on the accumulated chat records and task completion information. In the case of a \texttt{pause \& trigger} message, the framework prompts the client to generate the task IDs that require triggers and broadcasts them to all group members. For \texttt{discussion} or \texttt{task assignment} messages, they are directly broadcast to all participants in the group chat.

Upon receiving a message, the client parses it according to the agent message protocol. If the \texttt{next\_speaker} field does not include the current client, the message is simply added to the group chat history. However, if the client is designated as the next speaker, it must take appropriate actions based on the message type. For \texttt{discussion} messages, the client generates a response to continue the conversation. In the case of \texttt{sync} or \texttt{async task assignment} messages, the client extracts its assigned task from the chat record, summarizes it, and specifies the relevant information to be passed to the integrated agent. The agent then executes the task based on the summarized description and relevant chat messages, returning the result upon completion. If the message type is \texttt{pause \& trigger}, the client updates the corresponding task triggers in the Task Management Module.

The Communication Module, in conjunction with the other modules in the Interaction Layer and Data Layer, enables seamless and structured collaboration among agents. By adhering to the well-defined agent message protocol and leveraging the functionalities provided by the various modules, clients can effectively participate in discussions, assign tasks, and coordinate their actions to achieve the desired goals.

\subsection{Server}
\label{sec:appendix-ioa-detail-server}

The server component of \framework serves as the central hub for agent coordination, communication, and management. It comprises three layers: the Foundation Layer, the Data Layer, and the Interaction Layer. Each layer contains modules that work together to facilitate agent registration, discovery, and message routing. In this subsection, we provide a detailed description of the implementation of each module within the server's layers.

\subsubsection{Foundation Layer}
\label{sec:appendix-ioa-detail-server-foundation}

\paragraph{Network Infrastructure Module and Data Infrastructure Module}
The Network Infrastructure Module and Data Infrastructure Module in the server are largely similar to their counterparts in the client. However, the server's Data Infrastructure Module incorporates the use of the Milvus vector database to support the construction and maintenance of the Agent Registry. Milvus enables efficient similarity search and retrieval of agent information based on their characteristics, allowing the server to provide clients with the functionality to discover and match agents effectively.

\paragraph{Security Module}
While the Security Module is not extensively utilized in the current implementation of \framework, we acknowledge its crucial role in ensuring the integrity and reliability of the framework in real-world deployments. This module is responsible for verifying and controlling the integration of third-party agents into the clients, preventing malicious agents from compromising the entire framework. As \framework evolves, the Security Module will be enhanced to provide robust authentication, authorization, and monitoring mechanisms, safeguarding the collaborative environment from potential security threats.

\subsubsection{Data Layer}
\label{sec:appendix-ioa-detail-server-data}

\paragraph{Agent Registry Module}
The Agent Registry Module maintains a comprehensive record of all clients integrated into the server. When a client connects to the server, it is required to provide a detailed description of the integrated agent, including its name and capability description. This information is stored in the Agent Registry, enabling similarity matching based on agent characteristics. The Agent Registry serves as a central repository for agent information, facilitating agent discovery and team formation processes.

\paragraph{Session Management Module}
The Session Management Module is responsible for managing the WebSocket connections of all online agents and keeping track of the group chats they participate in. It maintains a mapping between agents and their respective WebSocket connections, as well as the associations between agents and group chats. When a client sends a message, the Session Management Module ensures that the message is properly routed to all clients involved in the corresponding group chat, guaranteeing reliable and efficient communication within the collaborative environment.

\subsubsection{Interaction Layer}
\label{sec:appendix-ioa-detail-server-interaction}

\paragraph{Agent Query Module}
The Agent Query Module handles incoming requests from clients seeking to discover and match agents based on specific characteristics. Upon receiving a query request, the module converts the provided characteristics into vector representations and performs similarity matching against the agents stored in the Agent Registry. The implementation of this module can vary depending on the specific requirements and scalability needs of the framework. For instance, techniques such as BM25 or other information retrieval methods can be employed to enhance the matching process and improve the relevance of the returned agent results.

\paragraph{Group Setup Module}
The Group Setup Module is responsible for handling client requests to create new group chats. When a client submits a request to set up a group chat, specifying the desired team members, the Group Setup Module processes the request and initializes a new group chat instance. It assigns a unique \texttt{comm\_id} to the newly created group chat and notifies all participating clients about their inclusion in the chat. The Group Setup Module works in conjunction with the Session Management Module to ensure that the necessary WebSocket connections and mappings are established for efficient communication within the group chat.

\paragraph{Message Routing Module}
The Message Routing Module plays a critical role in facilitating communication between clients within group chats. When a client sends a message, the Message Routing Module receives the message and parses it according to the agent message protocol. Based on the \texttt{comm\_id} specified in the message, the module identifies the corresponding group chat and forwards the message to all clients associated with that chat. The Message Routing Module leverages the information maintained by the Session Management Module to ensure accurate and timely delivery of messages to the intended recipients.

The server component of \framework, with its carefully designed modules and interactions, provides a robust and efficient infrastructure for agent coordination, communication, and management. By leveraging the capabilities of the Foundation Layer, Data Layer, and Interaction Layer, the server enables seamless agent discovery, team formation, and message exchange, fostering a collaborative environment where diverse agents can work together to achieve common goals.

As \framework continues to evolve, the server component will be further enhanced to incorporate advanced features such as load balancing, fault tolerance, and scalability, ensuring that the framework can handle the growing demands of real-world multi-agent systems. Additionally, the Security Module will be strengthened to provide comprehensive security measures, safeguarding the integrity and confidentiality of agent interactions within the framework.

\subsection{Implementation Details of Different Experiments}
\label{sec:appendix-exp-detail}

In this section, we provide an overview of the implementation details for each experiment conducted to evaluate the performance of \framework.

\subsubsection{GAIA}
\label{sec:appendix-exp-detail-gaia}

For the GAIA benchmark, \framework integrated four ReAct agents: Web Browser, Code Executor, YouTube Transcript Downloader, and Wikidata Searcher. The tools provided to Web Browser and Code Executor agents are adapted from the AutoGen framework with minor modifications to ensure compatibility with \framework. 
To address the YouTube-related tasks in GAIA, we develop a YouTube video transcript downloader based on PyTube\footnote{\url{https://github.com/pytube/pytube}}. For videos without readily available transcripts, the tool employs the Whisper model to transcribe spoken language into text. 
Similarly, we adapt the Wikidata tool from Langchain\footnote{\url{https://python.langchain.com/v0.1/docs/integrations/tools/wikidata/}} to fit the \framework ecosystem. These adaptations showcases a key feature of \framework: when a task requires a specific tool, it can be easily integrated into the system through its implementation and agent adaptation, enabling it to participate in task completion.

Due to budget constraints, we conduct performance testing on the GAIA validation set. Despite this limitation, the results provide valuable insights into the effectiveness of \framework in handling complex, multi-step tasks.

\begin{figure}
    \centering
    \includegraphics[width=\linewidth]{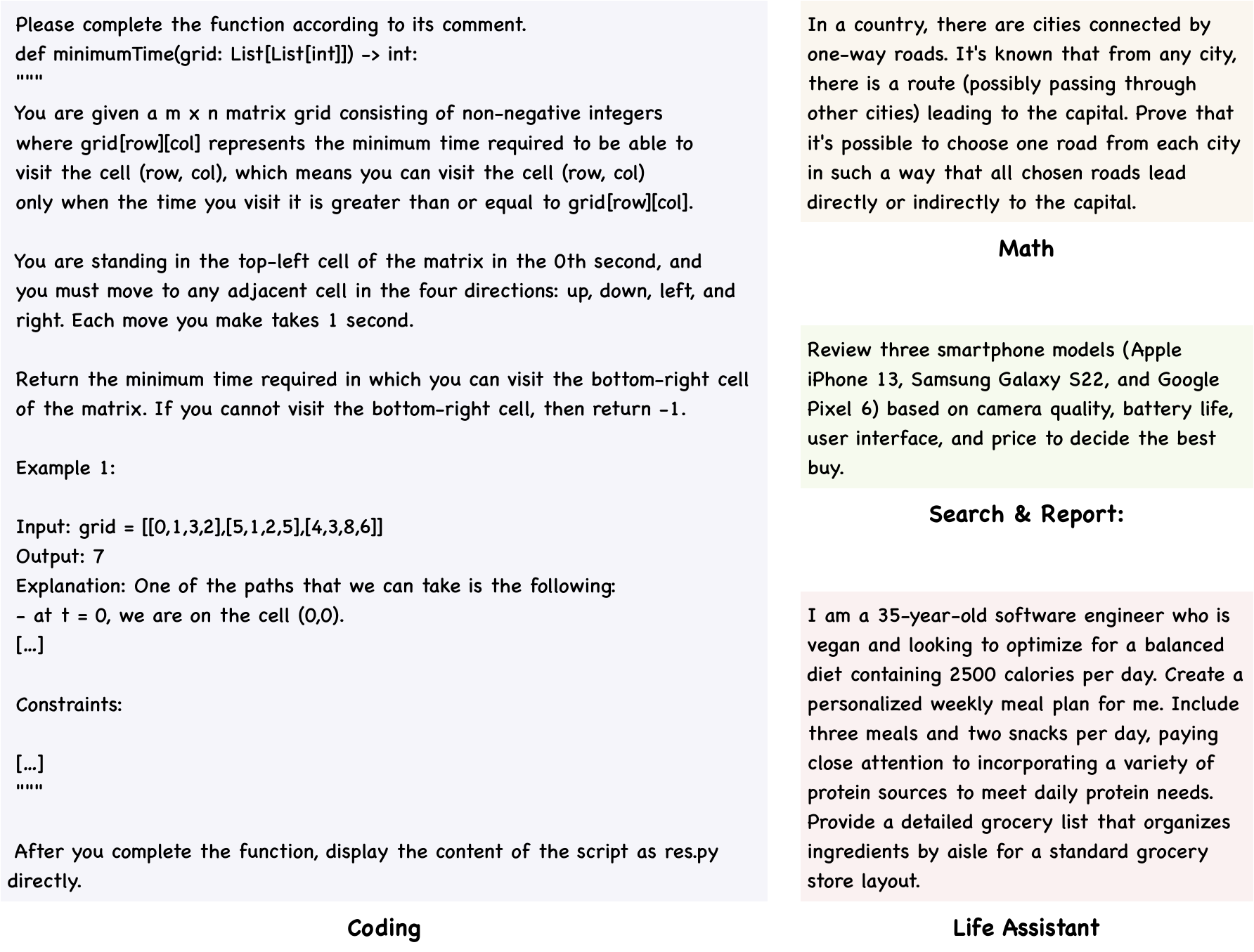}
    \caption{Example instructions from different categories in our open-ended instruction benchmark}
    \label{fig:open-ended-instruction-example}
\end{figure}
\subsubsection{Open-Ended Instruction Benchmark}
\label{sec:appendix-exp-detail-open-ended}
To create a diverse and challenging benchmark for evaluating the performance of \framework on open-ended tasks, we construct a set of 153 instructions spanning four categories: search \& report, coding, math, and life assistance. The benchmark construction process involved three main steps:

First, we select the instructions based on the real-world complex tasks used by XAgent~\citep{xagent2023}. These instructions were categorized into the four aforementioned groups. Second, to increase the diversity of the benchmark, we manually create an additional 10 complex tasks. Finally, we use the Self-Instruct method~\citep{DBLP:conf/acl/WangKMLSKH23} to generate approximately 200 instructions, using the previously selected instructions as seeds. After manual screening and modification, we obtained the additional 94 instructions, resulting in a total of 153 tasks. The benchmark eventually consists of 52 search \& report tasks, 30 coding tasks, 30 math tasks, and 41 life assistance tasks. By incorporating a diverse set of open-ended instructions, this benchmark allows for a comprehensive evaluation of the performance and versatility of \framework in handling a wide range of real-world scenarios. We show one example instruction for each category in~\cref{fig:open-ended-instruction-example}.

\textbf{Evaluation Methodology.}
For \framework, we consider the final conclusion generated by the agents as the final answer. However, since AutoGPT~\citep{Significant_Gravitas_AutoGPT} and Open Interpreter~\citep{OpenInterpreter} complete tasks in multiple steps and do not inherently generate a conclusion, we prompted them to provide a detailed conclusion as the final answer after task completion.

Inspired by the pairwise comparison evaluation method used in MT-Bench~\citep{zheng2023judging}, we employ GPT-4 to evaluate the responses of \framework against AutoGPT and Open Interpreter. To mitigate potential biases introduced by the order of the responses, we alternate the order of the two responses when presenting them to GPT-4 for evaluation. A result is counted as a \textit{win} for a system only when it is consistently determined to be superior to its competitor in both orderings. In cases where the performance is inconsistent across the two orderings, the result is considered a \textit{draw}.

\subsubsection{Embodied Agent Tasks}
\label{sec:appendix-exp-detail-embodied}

For the RocoBench experiments, we adhere to the original paper's methodology, which relies on discussions and parsing specific formatted strings from the discussion results to determine the embodied agent's actions, rather than using agents to call tools directly. We implement two clients that communicate without integrated agents, requiring them to output strings in the RocoBench format at the conclusion stage. These strings are then parsed and used to interact with the environment using RocoBench's predefined parsing functions. This approach serves as a validation of \framework's client implementation and communication mechanism design.

To accommodate the varying requirements of different tasks in RocoBench, we adopt task-specific settings. For the Sort, Sandwich, and Sweep tasks, which exhibit strong interdependencies between steps, we retained the chat history and continued each new action discussion based on the previous group chat. In contrast, for the Cabinet and Rope tasks, where the steps were less interdependent, we initiated a new group chat for each action to optimize costs. Other settings remained consistent with the Roco Dialog baseline.

\subsubsection{Retrieval-Augmented Generation}
\label{sec:appendix-exp-detail-rag}

For the retrieval-augmented generation (RAG) question-answering task, we follow the settings outlined in Apollo's Oracle. We provide agents with two evidence pools: one derived from Wikipedia and the other from Google. For Wikipedia, we utilize Pyserini's pre-built index of Wikipedia content up to January 20, 2021, retrieving the top 10 most relevant results for each query. For Google, we directly access the Google Search API, returning the top 5 most relevant results for each query. These tools were made available to the client-side LLMs, enabling them to query relevant information during discussions and ultimately provide well-informed answers.

To evaluate the performance of \framework on the RAG task, we randomly sample 500 entries from the validation or test sets of the four datasets. After the model generates answers, we employ GPT-4 for answer evaluation. Specifically, we provide GPT-4 with the dataset answers and the model's answers, requiring it to output its reasoning in a Chain of Thought (CoT) manner before providing a final correctness judgment. 

\section{Visualization of RocoBench}
\label{sec:appendix-rocobench-visualization}
\begin{figure}
    \centering
    \begin{subfigure}[b]{0.19\textwidth}
        \includegraphics[width=\textwidth]{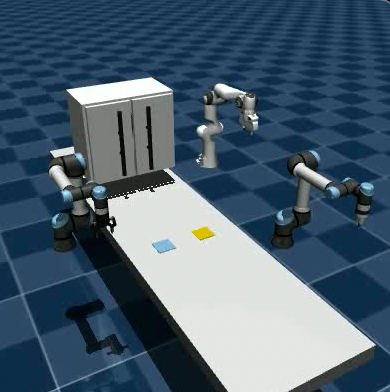}
        \caption{Cabinet}
    \end{subfigure}
    \begin{subfigure}[b]{0.19\textwidth}
        \includegraphics[width=\textwidth]{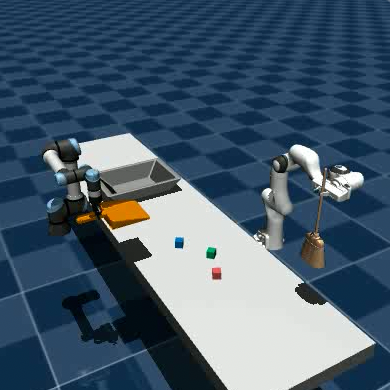}
        \caption{Sweep}
    \end{subfigure}
    \begin{subfigure}[b]{0.19\textwidth}
        \includegraphics[width=\textwidth]{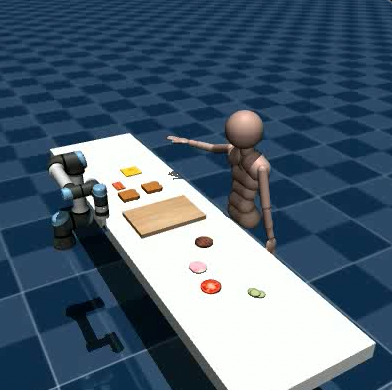}
        \caption{Sandwich}
    \end{subfigure}
    \begin{subfigure}[b]{0.19\textwidth}
        \includegraphics[width=\textwidth]{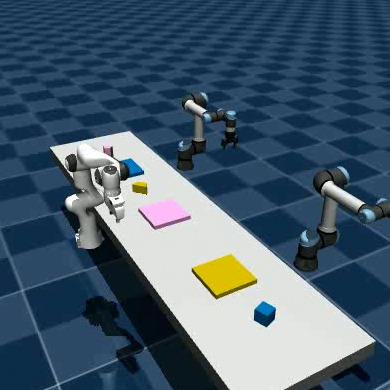}
        \caption{Sort}
    \end{subfigure}
    \begin{subfigure}[b]{0.19\textwidth}
        \includegraphics[width=\textwidth]{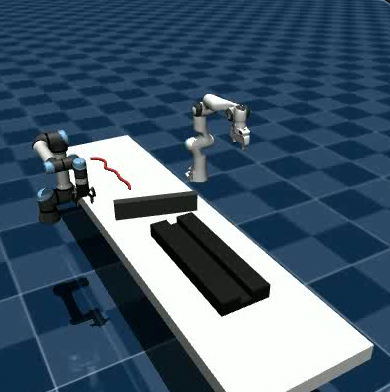}
        \caption{Rope}
    \end{subfigure}
    \caption{The different environments in RocoBench.}
    \label{fig:roco}
\end{figure}
We provide the visualization of RocoBench at~\cref{fig:roco}. The \textbf{cabinet task} requires three agents to collaborate: two agents open and hold the cabinet door while the third agent retrieves two cups from inside the cabinet and places them onto coasters that match the color of the cups. The \textbf{sweep task} involves two agents coordinating their actions: one agent controls a broom to sweep cubes, while the other agent holds a bucket to collect the cubes, and finally, they dump all the cubes into a dustbin. In the \textbf{sandwich task}, two agents work together to pick up ingredients and stack them according to a given recipe. The \textbf{sort task} requires three agents to place three cubes onto coasters with matching colors. Since each agent can only reach a limited area, they must coordinate their movements. Lastly, the \textbf{rope task} involves agents moving a rope into a bracket. They must communicate effectively to decide the correct path for maneuvering the rope.

\section{Simulated Environment for Team Formation Evaluation}
\label{sec:appendix-teamup-detail}

\subsection{Regular Team Formation simulated environment Construction}
\label{sec:appendix-regular-team-formation}
To construct a simulated environment for evaluating the regular ,team formation mechanism, we employ GPT-4-1106-preview to generate a diverse set of tasks and agents. The dataset construction process involved the following steps:

\begin{enumerate}
    \item Task Generation:
    \begin{itemize}
        \item Using ChatGPT-4, we generate 399 distinct categories of theme keywords, covering various domains such as sports, lifestyle, and entertainment.
        \item From these categories, we randomly select 25 themes and task GPT-4 with generating task descriptions related to at least four themes from the selected set, thus obtaining a task that require diverse agents with different capabilities.
        \item Task descriptions are generated in JSON format using the GPT-4 API, ensuring a structured and consistent representation.
    \end{itemize}
    \item Agent Generation:
    \begin{itemize}
        \item After generating the tasks, for each task, we again prompt GPT-4 to construct at least two agents with varying capabilities for the given task, including the name of the agent, the type of the agent and the description of the agent.
        \item The agent profile format is designed to align with the server-side agent registry, facilitating seamless integration and interaction within \framework.
    \end{itemize}
\end{enumerate}

An example of a generated task description in JSON format is as follows:

\begin{lstlisting}[language=json]
{
  "task_id": "xxx",
  "task_description": "Develop a mobile app that helps users plan and manage their personal finance, including budgeting, expense tracking, and investment suggestions."
}
\end{lstlisting}

Similarly, an example of an agent profile in JSON format is:

\begin{lstlisting}[language=json]
{
  "agent_name": "FinanceGuru",
  "agent_type": "Thing Assitant"
  "agent_description": "FinanceGuru is a highly skilled agent specializing in personal finance management. It has extensive knowledge of budgeting techniques, expense tracking tools, and investment strategies. FinanceGuru can provide personalized recommendations based on a user's financial goals and risk tolerance."
}
\end{lstlisting}

A complete example with agent profiles and task description in JSON format is:
\begin{lstlisting}[language=json]
{
  "agents": [
   {
     "agent_name": "BeautyRoutineAssistant",
     "agent_type": "Thing Assistant",
     "agent_description": "This agent specializes in grooming and beauty routines. It is designed to offer personalized beauty tips and tutorials for efficient makeup application based on the user's facial features, skin type, and preferences. It suggests makeup looks that align with weather conditions and the user's daily agenda. The assistant can interface with smart mirrors, makeup organizers, and tutorials for a streamlined morning routine."
   },
   {
     "agent_name": "LanguageCoachAssistant",
     "agent_type": "Human Assistant",
     "agent_description": "This is an educational aide focused on facilitating language learning sessions. It assesses the user's current language proficiency, learning style, and daily schedule to allocate an optimal one-hour learning window. The agent customizes lesson plans, integrates with language learning apps or platforms, and can organize virtual interactions with native speakers for immersive learning experiences."
   },
   {
     "agent_name": "EcoCuisineAssistant",
     "agent_type": "Thing Assistant",
     "agent_description": "EcoCuisineAssistant is dedicated to healthy meal planning and environmental consciousness. It suggests simple, nutritious dinner recipes based on dietary needs, kitchen inventory, and prep time constraints. It interfaces with smart kitchen appliances to guide the cooking process and monitors waste to teach and reinforce correct recycling habits, ensuring a minimized environmental impact."
   } 
  ],
  "task_description": "I am looking to create a daily routine that incorporates applying makeup efficiently in the morning, spending an hour learning a new language, preparing a simple and healthy dinner, and correctly recycling the waste generated throughout the day."
}
\end{lstlisting}


\subsection{Nested Team Formation simulated environment Construction}
\label{sec:appendix-nested-team-formation}
In a similarly way, in order to construct a simulated environment for evaluating the nested team formation mechanism, we also employ GPT-4-1106-preview to generate two diverse sets of tasks and agents. The dataset construction process involved the following steps:

\begin{enumerate}
    \item Sub-tasks Completed by Existing Agents:
    \begin{itemize}
        \item Su-btask Generation:
        \begin{itemize}
            \item Based on the dataset that we have constructed for regular team formation, we randomly select 300 sets as the original dataset.
            \item For tasks in the original dataset, we prompt GPT-4 to construct a sub-task that can be completed by an existing agent, with the agent being selected by GPT-4.
            \item Sub-task description are generated in JSON format using the GPT-4 API with the existing agent, ensuring a structured and consistent representation.
        \end{itemize}
            
    \end{itemize}
    \item Sub-tasks Completed by Additional Agent:
    \begin{itemize}
        \item Sub-task and Agent Generation:
        \begin{itemize}
            \item After generating the sub-tasks for exiting agent, we take the rest of sets as the another original dataset.
            \item The difference for sub-task completed by existing agent is that we prompt GPT-4 to construct a sub-task requiring a very specific expertise. 
            \item Meanwhile, we also prompt GPT-4 to construct an agent with distinct capabilities compared to the existing agents to complete the generated sub-task, including the name of the agent, the type of the agent and the description of the agent.
            \item Sub-task description and additional agent are generated in JSON format using the GPT-4 API ensuring a structured and consistent representation.
        \end{itemize}
    \end{itemize}
\end{enumerate}

An example of a generated sub-task description with existing agent in JSON format is as follows:

\begin{lstlisting}[language=json]
{
  "additional_subtask": {
  "task_description": "Develop a comprehensive marketing plan highlighting the business's commitment to sustainability, including strategies for podcast promotion, brand awareness, and customer engagement.",
  "agent": {
      "agent_name": "MarketingStrategist",
      "agent_type": "Human Assistant",
      "agent_description": "Critical to the success of the sustainability-focused business, this agent is in charge of advertising campaigns, social media presence, and public relations. With a strong emphasis on the company's eco-friendly values, it develops targeted marketing strategies to reach a wider audience, creating a strong brand identity around sustainability. The agent also handles analytics, gauging the effectiveness of marketing efforts and adjusting tactics to optimize outreach and customer engagement."
  },
  "agents": [
  {
      "agent_name": "SustainabilityEducator",
      "agent_type": "Human Assistant",
      "agent_description": "This agent is specialized in creating, curating, and disseminating information about sustainable living. It is responsible for researching various subjects related to sustainability, structuring podcast content, interviewing experts, and sharing practical tips on incorporating eco-friendly practices into daily life. The agent will also engage the audience through various channels, answer listener queries, and promote discussion on sustainability."
  },
  {
      "agent_name": "EcoDesigner",
      "agent_type": "Human Assistant",
      "agent_description": "Tasked with the creation of custom eco-friendly products, this agent has expertise in sustainable design practices and materials. It collaborates with customers to understand their needs and preferences, and uses innovative methods to craft personalized, environmentally responsible goods while maintaining aesthetic and functional standards. Additionally, the agent works closely with suppliers to ensure the sustainability and ethical sourcing of raw materials."
  },
  {
      "agent_name": "MarketingStrategist",
      "agent_type": "Human Assistant",
      "agent_description": "Critical to the success of the sustainability-focused business, this agent is in charge of advertising campaigns, social media presence, and public relations. With a strong emphasis on the company's eco-friendly values, it develops targeted marketing strategies to reach a wider audience, creating a strong brand identity around sustainability. The agent also handles analytics, gauging the effectiveness of marketing efforts and adjusting tactics to optimize outreach and customer engagement."
      }
  ],
"task_description": "I want to start a business that focuses on sustainable living. The business will include a podcast series on how to incorporate sustainability into daily life and crafting custom eco-friendly products for customers."
}
\end{lstlisting}

Similarly, an example of a generated sub-task description with additional agent in JSON format is:

\begin{lstlisting}[language=json]
{
  "additional_subtask": {
  "task_description": "Implement advanced custom animations and interactive elements to enhance the visual appeal of the personal website, particularly for the graphic design portfolio section. This includes creating dynamic, engaging animations that showcase the artist's skills and bring the homepage to life, as well as ensuring cross-browser compatibility and responsiveness on various devices.",
  "agent": {
    "agent_name": "AnimationExpert",
    "agent_type": "Thing Assistant",
    "agent_description": "AnimationExpert is a highly specialized virtual assistant dedicated to creating sophisticated web animations and interactive experiences. It is equipped with state-of-the-art tools and knowledge of the latest animation libraries like GSAP, Three.js, and WebGL. This agent analyzes the existing style and content of the website to develop tailored, eye-catching animations that complement the graphical elements without compromising website performance. It ensures compatibility with all major browsers and devices and works seamlessly with responsive design principles to deliver a consistent experience across all user interfaces."
    }
  },
  "agents": [
  {
    "agent_name": "WebDesignerAssistant",
    "agent_type": "Human Assistant",
    "agent_description": "This agent specializes in web design and user experience. It assists in creating a visually appealing and intuitive homepage layout that effectively showcases the portfolio of graphic design work. It will help organize content in a cohesive manner, using best web design practices to emphasize the most compelling pieces. This assistant can also suggest and implement design elements that reflect personal style and artistic sensibility."
  },
  {
    "agent_name": "ContentStrategistAssistant",
    "agent_type": "Human Assistant",
    "agent_description": "This agent focuses on content creation and management. It supports in putting together the fashion and style blog posts by helping to curate topics, edit posts for clarity and brand consistency, and integrate them into the website. It ensures that the blog content is strategically placed for optimal engagement, incorporating SEO best practices to increase visibility and draw in more visitors interested in fashion and style."
  },
  {
    "agent_name": "PhotographyShowcaseAssistant",
    "agent_type": "Thing Assistant",
    "agent_description": "This agent is tailored to enhance the presentation of photography work on the website. Equipped with image organizing and editing software integration capabilities, it can help sort and select the best photographs to feature. It will ensure that the images are displayed in high quality and that the loading speed is optimized for user convenience. This assistant will also provide options for interactive image galleries that enable visitors to view the work in detail."
  }
],
  "task_description": "I want to create a personal website that showcases my portfolio of graphic design work, my fashion and style blog posts, and my photography. Please provide instructions on how to design the layout for my homepage that effectively incorporates all three aspects."
}
\end{lstlisting}

By generating a couple of diverse sets of tasks and agents, we create a comprehensive simulated environment for evaluating the regular team formation mechanism and the nested team formation mechanism. This environment enables us to assess the effectiveness of \framework in assembling appropriate teams to complete task requirements, addressing the limitations of existing benchmarks in providing suitable large-scale agent evaluation scenarios.

\end{document}